%% file: main.tex
\documentclass[10pt,twocolumn,letterpaper]{article}

\usepackage[pagenumbers]{cvpr}
\input{preamble}

\definecolor{cvprblue}{rgb}{0.21,0.49,0.74}
\usepackage[pagebackref,breaklinks,colorlinks,allcolors=cvprblue]{hyperref}

\title{CASA: Cross-Attention over Self-Attention for Efficient Vision-Language Fusion
\vspace{-.3cm}
}

\author{
    \begin{tabular}{ccccc}
       Moritz Böhle$^*$ & Amélie Royer$^*$ & Juliette Marrie$^*$ & Edouard Grave & Patrick Pérez \\
    \small{moritz@kyutai.org} & \small{amelie@kyutai.org} & \small{juliette@kyutai.org} & \small{egrave@kyutai.org} & \small{patrick@kyutai.org}
    \end{tabular}
}

\begin{document}
\maketitle
\boldmaintabletrue
\input{sec/0_abstract}    
\input{sec/1_intro}
\input{sec/2_related}
\input{sec/3_method}
\input{sec/4_experiments}

\input{sec/conclusion}
\clearpage

\myparagraph{Acknowledgements.} 
This project is funded by Iliad Group, CMA CGM Group and Schmidt Sciences. The authors thank Alexandre D\'{e}fossez for his support and feedback throughout the project. 

{
    \small
    \bibliographystyle{ieeenat_fullname}
    \bibliography{main}
}
\clearpage
\input{sec/X_suppl}
\end{document}

%% file: preamble.tex
\usepackage{pgffor}
\usepackage[most]{tcolorbox}
\usepackage{xcolor}
\usepackage[normalem]{ulem}
\usepackage{fontawesome5}
\usepackage{dsfont}
\usepackage{pifont}
\usepackage{amsmath}
\usepackage{amssymb}
\usepackage{bbding}
\usepackage{booktabs}
\usepackage[utf8]{inputenc}
\usepackage[english]{babel}
\usepackage{csquotes}
\usepackage{makecell}
\usepackage{subcaption}
\usepackage{tikz}
\usepackage{graphicx}
\usepackage{multirow}
\usepackage{adjustbox}
\newcommand{\cmark}{\ding{51}}%
\newcommand{\xmark}{\ding{55}}%
\usepackage{spreadtab} 
\usepackage{siunitx}   
\usepackage{booktabs}
\usepackage{arydshln}

\newcommand{\myparagraph}[1]{\vspace{0.1cm}\noindent {\bf #1}}

\usepackage{xspace}

\newcommand{\CA}[1][]{CA$_{\text{\scriptsize #1}}$\xspace}

\newcommand{\myeq}{\mkern1.25mu{=}\mkern1.25mu}

\newcommand{\chartqa}{\sc{Chart}}
\newcommand{\docvqa}{\sc{Doc}}
\newcommand{\infovqa}{\sc{Info}}
\newcommand{\ocrbench}{\sc{OcrB}}
\newcommand{\textvqa}{\sc{Text}}
\newcommand{\realworldqa}{\sc{RealW}}
\newcommand{\aitd}{\sc{AI2D}}
\newcommand{\gqa}{\sc{GQA}}
\newcommand{\mme}{\sc{MME}}

\newif\ifboldmaintable

\newcommand{\ours}{CA\xspace}

\newcommand{\CAtv}{$\text{CA}_{\text{t}\!+\!\text{v}}$\xspace}

\newcommand{\CAtvl}{$\text{CA}_{\text{t}\!+\!\text{v}}^{\,\text{\tiny\faFeather}}$\xspace}
\newcommand{\CAs}{$\text{CA}^{\,\text{\tiny\faLink}}$\xspace}
\newcommand{\CAl}{$\text{CA}^{\,\text{\tiny\faFeather}}$\xspace}
\newcommand{\CAHe}{$\text{CA}_{\text{He-2B}}$\xspace}
\newcommand{\CAQw}{$\text{CA}_{\text{Q2.5-VL}}$\xspace}
\newcommand{\CAtvHe}{$\text{CA}_{\text{tv,\,He-2B}}$\xspace}
\newcommand{\CAtvQw}{$\text{CA}_{\text{tv,\,Q2.5-VL}}$\xspace}
\newcommand{\CAlHe}{$\text{CA}_{\text{He-2B}}^{\,\text{\tiny\faFeather}}$\xspace}

\newcommand{\CAsHe}{$\text{CA}_{\text{He-2B}}^{\,\text{\tiny\faLink}}$\xspace}

\newcommand{\timestamp}[1]{%
  \raisebox{0.25ex}{\textcolor{gray}{\tiny\textsf{\bf[#1]}}}%
}

\newenvironment{subtitlebox}{%
  \begin{tcolorbox}[enhanced, colback=gray!2, colframe=gray!40, boxrule=0.5pt, arc=2mm, left=4mm, right=4mm, top=2mm, bottom=2mm
  ]%
}{\end{tcolorbox}}

\newcommand{\rep}[1]{
\phantom{$^\dagger$}\;#1{\color[rgb]{.5, .5, .5}$^\dagger$}
}

\makeatletter

\newcommand{\suptableofcontents}{%
  \begingroup
    \let\saved@l@section\l@section
    \let\saved@l@subsection\l@subsection
    \let\saved@l@subsubsection\l@subsubsection

    \def\l@section##1##2{}%
    \def\l@subsection##1##2{}%
    \def\l@subsubsection##1##2{}%

    \def\l@suptocmarker##1##2{%
      \let\l@section\saved@l@section
      \let\l@subsection\saved@l@subsection
      \let\l@subsubsection\saved@l@subsubsection
    }%

    \tableofcontents
  \endgroup
}

\let\svthefootnote\thefootnote
\newcommand\freefootnote[1]{%
  \let\thefootnote\relax%
  \footnotetext{#1}%
  \let\thefootnote\svthefootnote%
}
\newcommand{\hcref}[1]{\hyperref[#1]{\cref{#1}}}
\newcommand{\hCref}[1]{\hyperref[#1]{\Cref{#1}}}

%% file: sec/0_abstract.tex
\begin{abstract}
Vision-language models (VLMs) are commonly trained by  directly inserting image tokens from a pretrained vision encoder into the text stream of a language model. This allows text and image information to fully attend to one another within the model, but becomes rapidly costly for  long multi-image conversations or streaming video applications, both in terms of memory and compute. VLMs leveraging cross-attention (CA) are an efficient alternative to token insertion as image tokens are not added to the KV cache. Despite being introduced early on, multimodal CA  models are scarce in the current VLM literature and often underperform their token insertion counterparts. 
In this work, we reinvestigate the effectiveness of cross-attention for vision-language modeling: (i) We analyze the core differences between the cross-attention and self-attention mechanisms, 
(ii) we train cross-attention VLMs both from a text-only LLM and by adapting a pretrained insertion-based VLM, showing that simple cross-attention is far more competitive with token insertion than previously reported,
and (iii) we demonstrate the practical advantages of cross-attention on real-time video captioning, where it naturally maintains low latency and near-constant memory cost.
For samples and code, please see our project page at \url{kyutai.org/casa}.
\freefootnote{$^*$\,Equal contribution.}
\end{abstract}

%% file: sec/1_intro.tex
\section{Introduction}
\label{sec:intro}
Cross-attention (CA) has been employed early on as a lightweight mechanism for fusing multimodal information in transformers~\citep{alayrac2022flamingo,li2023blip2,Tan2019LXMERTLC}.
Most recent state-of-the-art VLMs have departed from cross-attention-based fusion, and instead insert visual embeddings 
into the language model's input stream, directly interleaving them with the text embeddings~\citep{chen2024internvl25,bai2025qwen25vl,zhang2025videollama3,wang2024qwen2vl}.
While insertion-based fusion is remarkably effective, it incurs high computational and memory costs which grow with the number of image tokens, becoming a bottleneck for high-resolution images, multi-image conversations, and streaming video applications.

CA has recently regained interest as a naturally efficient alternative, particularly for streaming applications on long multimodal sequences~\cite{ye2025mplugowl3,chen2024evlm,royer2025moshivis,liu2024streamchat}.
This has led to multiple architectural variants of cross-attention, in an effort to improve its performance,  such as attention gating mechanisms~\citep{ye2025mplugowl3,royer2025moshivis}, learnable visual tokens~\citep{chen2024evlm}, or updating visual embeddings across the depth of the network~\citep{liu2024streamchat}.
Nonetheless, CA-based VLMs currently lag behind insertion-based models on several tasks such as document and chart understanding, as shown in \hyperref[fig:fig1]{Figure~\ref{fig:fig1}}.
The causes of this performance gap remain  poorly understood. In particular, it is unclear whether it reflects fundamental limitations of CA or instead arises from differences in training data  and implementation choices.
\input{figs/fig1}

Through this work, we methodically revisit CA as a fusion mechanism. To shed light on why current SotA CA models lag behind recent VLMs,  we first analyze the efficiency trade-offs of token insertion and cross-attention showing that  CA offers significant memory and speed advantages at both training and inference time (\hcref{tab:ca_vs_ins}). We validate these findings in two controlled settings: training a CA-based VLM from scratch from a text-only LLM, and adapting a pretrained insertion-based VLM to cross-attention. 
We find that vanilla cross-attention, without any bells and whistles, performs better than previously reported, narrowing the gap to token insertion to only a few percents on most benchmarks when compared in an identical training setting.
This simple model also outperforms current state-of-the-art CA-based VLMs of larger size, showing the importance of using a comparable modern training pipeline when assessing the performance of CA models with respect to the literature.  
We further evaluate our CA  model on the costly task of real-time video captioning, where CA maintains near-constant memory and latency over long video horizons.
In summary, we provide a controlled and comprehensive comparison of cross-attention and token insertion for vision-language fusion. Our contributions are threefold:
\textbf{\color[rgb]{.1, .5, .6}(i)} We systematically analyze the core differences between the two mechanisms, 
identifying {five} key core design elements that progressively bridge cross-attention and token insertion (\hCref{sec:differences}).
{\textbf{\color[rgb]{.1, .5, .6}(ii)} 
We train cross-attention VLMs both from a text-only LLM and by adapting a pretrained insertion-based VLM, showing that simple cross-attention is far more competitive with token insertion than previously reported.}
\textbf{\color[rgb]{.1, .5, .6}(iii)} We demonstrate the practical advantages of cross-attention on real-time video captioning~\cite{chen2025livecc}, where CA enables continuous visual updates with low latency and near-constant memory cost, while token insertion models quickly exhaust their memory budget.
To foster reproducibility, we release our inference code and trained models.

%% file: figs/fig1.tex
\begin{figure*}[t]
    \centering
    \includegraphics[width=0.95\linewidth]{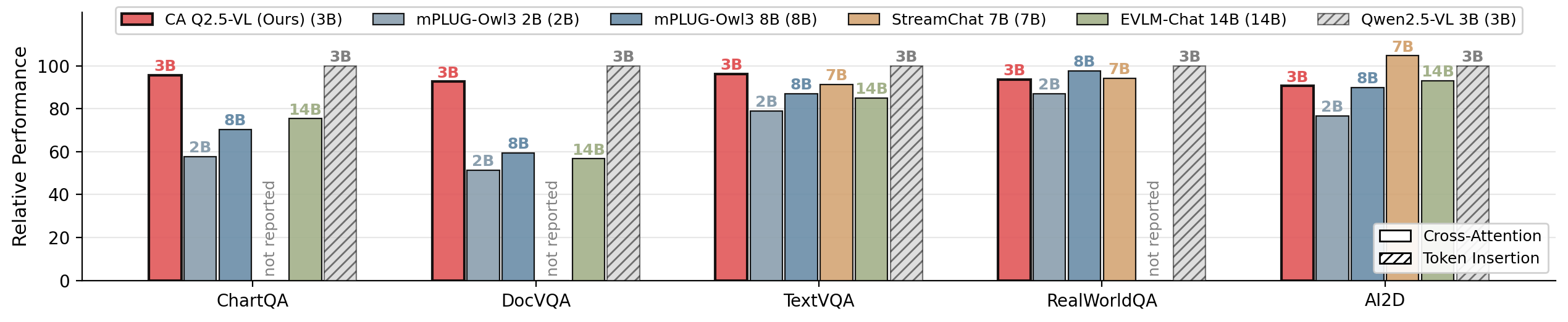}
    \vspace{-.1cm}
    \caption{\textbf{State-of-the-art cross-attention (CA) VLMs.} We report benchmark performance  normalized by Qwen2.5-VL 3B scores {\color[rgb]{.6, .6, .6}{(\textbf{hatched grey})}}. 
    We adapt Qwen2.5-VL 3B, a token-insertion-based VLM, to use cross-attention layers. The resulting model {\color[RGB]{202, 114, 108}(\textbf{red})} retains most of the base model's performance, and outperforms prior CA-based VLMs across model scales (2B--14B), in particular on high-resolution document and chart understanding tasks (ChartQA, DocVQA).
    \vspace{-.2cm}
    }
    \label{fig:fig1}
\end{figure*}

%% file: sec/2_related.tex
\section{Related work}
\label{sec:related}

\myparagraph{Insertion-based fusion.}
Token insertion has become the dominant paradigm for training  VLMs. For this, visual embeddings from a pretrained encoder are inserted directly into the language model's input sequence, interleaving them with text embeddings. The visual and textual information  thus interact through self-attention layers without any architectural changes.
Recent token insertion models~\citep{wang2024qwen2vl,bai2025qwen25vl,chen2024internvl25,zhang2023videollama,marafioti2025smolvlm} have  achieved strong multimodal performance with this strategy.
However, the number of visual tokens rapidly grows with image resolution or video length, thus increasing the memory and computational costs at both training and inference time.
A wide range of techniques has been explored to mitigate this cost, including compressing visual tokens via pixel unshuffling~\citep{shi2016real}, token merging~\citep{li2024videochatflash}, query-based compression~\citep{li2023blip2,zhang2023videollama,li2024mvbench}, or pooling~\citep{wang2025internvideo2,maaz2023videochatgpt}, inserting visual tokens in only a subset of layers~\citep{chen2024fastv}, reducing the cost of attention operations~\citep{Yang2024ParallelizingLT}, compressing the KV cache at inference~\citep{qian2024videostreaming,chen2025livecc,zhang2025flashvstream}, or modifying positional embeddings to handle longer contexts~\citep{Ge2024V2PEIM,bai2025qwen25vl}.
These techniques are largely orthogonal to the choice of fusion mechanism, and could also complement CA to further reduce the number of keys and values in the CA layers. 

\paragraph{Cross-attention-based fusion}
for VLMs was popularized by Flamingo~\citep{alayrac2022flamingo}, one of the first large-scale VLMs, in which a frozen LLM is conditioned on visual inputs through gated cross-attention.
It was adopted in subsequent works~\citep{awadalla2023openflamingo,li2025otter}, and revisited in \citep{ye2025mplugowl3,chen2024evlm,liu2024streamchat}, which leverage the natural scalability of cross-attention for long-context or streaming applications.
However, current SotA cross-attention VLMs  underperform token-insertion models, in particular on tasks requiring fine-grained visual understanding such as chart or document reading~\citep{ye2025mplugowl3,liu2024streamchat}.
Prior work has attempted to address this gap by, \eg, updating the visual representations throughout the depth of the model~\citep{liu2024streamchat}, adding register tokens~\citep{chen2024evlm}, or introducing bespoke architectural changes~\citep{ye2025mplugowl3}.
In this work, we show that the performance gap is much lesser than expected when comparing token-insertion to cross-attention fusion in identical pipelines, even when using vanilla cross-attention. 

\paragraph{Streaming visual understanding.}
Recently, the VLM literature has seen a gain of interest for streaming video understanding tasks, such as live video captioning or multimodal assistants~\cite{chen2025livecc,zhang2025flashvstream,xu2025streamingvlm}.
Such tasks imply a strict memory and computational budget to execute in real-time and for as long as possible.
To address the limitations of token insertion, it is possible to limit the KV cache size through compression~\citep{zhang2025flashvstream}, pruning of the oldest video frames~\citep{qian2024videostreaming,chen2025livecc}, or by adding condensed ``visual memories'' to the text stream~\citep{xu2025streamingvlm,zhang2024internlm}.
In contrast, cross-attention naturally lends itself to efficiently handle a dense stream of images as input.
Closest to our work, StreamChat~\citep{liu2024streamchat} adopts a cross-attention-based design that updates its visual keys and values at each decoding step to align the current text stream with the latest video frame.
However, to improve the performance of its cross-attention mechanism, StreamChat applies additional dedicated FFN layers to the visual tokens throughout the network.
In our ablations, we show that the additional updates of image embeddings indeed yield a small performance boost, but incur high memory and compute cost, in particular during training.

%% file: sec/3_method.tex
\section{To Cross-Attend or Self-Attend?}

\label{sec:method}

Cross-attention's practical benefits stem from the core idea that image tokens never directly enter the transformer's input token stream. Consequently, they are neither added to the KV cache, nor 
updated via the Feed Forward Network (FFN) layers, saving both compute and memory. 
However, these strengths may become a limitation when considering the model's performance: Unlike token insertion, image embeddings are not updated throughout the depth of the network, and text tokens cannot directly attend to past images as they are not present in the KV cache. 
To better understand these distinctions and how they  impact the performance-efficiency trade-off, we first formalize  the CA and self-attention (SA) mechanisms (\hCref{sec:preliminaries}) and analyze their key differences (\hCref{sec:differences}). Finally, we discuss how to  implement CA for scalable multi-image training, allowing us to leverage sequence packing strategies and video inputs (\hCref{sec:efficient-impl}).

\subsection{Preliminaries}
\label{sec:preliminaries}
\myparagraph{Self-attention in token insertion models.}
Given text tokens $x = x_{1 \dots T}$ and a tokenized image $y = y_{1 \dots N}$ inserted at position $K<T$ in the text stream, token insertion lets $x_T$ interact with image tokens through self-attention  in a standard causal setting as:
\begin{align}
    \textbf{SA}_{\text{ins}}(x_T | x_{i\leq T}, y) = \text{MHA}(x_T |\, x_{1\dots K}\, y_{1\dots N}\, x_{K+1\dots T})\,.
    \label{eq:token_insertion}
\end{align}
$\text{MHA}(q| k)$ is multi-head attention~\citep{vaswani2017attention} with query $q$ and $k$ as keys and values.

\myparagraph{Cross-attention.}
In contrast, CA injects visual information to text tokens as additive updates. In CA layers, text tokens $x_{i > K}$ which follow the image at timestep $K$ attend to the corresponding image tokens as keys and values:
\begin{align}
    \textbf{CA}(x_T | y) = \mathds{1}_{T>K} \times \text{MHA}(x_T|\, y_{1\dots N}) \,.
    \label{eq:cross_attention}
\end{align}
In the remainder of the text, we will refer to the interval $[K, T]$ 
as the ``window'' corresponding to the image $y$ seen at time $K$. Simply speaking, each window is defined as the temporal positions between two input images. Note that in this setting, text tokens $x_{i > K}$ do not attend to past images seen before $y$ by design, and we will further elaborate on this locality property in \hCref{sec:differences}.
The output of \eqref{eq:cross_attention} is then combined, typically via a sum, with the output of the corresponding SA layer in the same transformer block, which operates on text tokens only: 
\begin{align}
    \textbf{SA}_{\text{text}}(x_T |\, x_{i\leq T}) = \text{MHA}(x_T| \, x_{1\dots T}).
    \label{eq:sa_text}
\end{align}
In practice, we only consider the scenario where cross-attention layers are placed \textit{in parallel} of self-attention layers in the same transformer block, as illustrated in \hCref{subfig:arch}: Both CA and $\text{SA}_{\text{text}}$ operate as separate layers, with their own projections, but on the same inputs. Nevertheless, other designs have been explored in the literature, such as placing them after \citep{liu2024streamchat}. 
The attention patterns of
\hcref{eq:token_insertion}, \hcref{eq:cross_attention}, and \hcref{eq:sa_text} are summarized in \hCref{fig:attention_pattern}. 

\subsection{From Cross-Attention to Token Insertion}
\label{sec:differences}

We identify {five} core differences between cross-attention and token insertion as fusion mechanisms, each with a different impact on model efficiency and performance. 
We discuss each of them below, emphasizing how, by combining these core aspects, we can progressively recover self-attention from cross-attention. 

\input{figs/attention_pattern}
\input{tables/table1}

\myparagraph{(D1) Additional parameters.}
CA introduces new dedicated layers with additional learnable parameters.
In contrast, SA uses the same projection weights to process image and text tokens without distinction.
To eliminate this difference, we introduce a  parameter-sharing variant of CA, denoted by \CAs, in which the query, key, value and output projections are shared between the CA and SA layers of a same transformer block.
Like token insertion, this variant thus adds no new parameters. Further, since CA and $\text{SA}_{\text{text}}$ operate in parallel on the same inputs, we only need to compute said projections once, thereby saving compute.

\myparagraph{(D2) Joint text-image attention and positional embeddings.}
Inserting images into the token stream allows text tokens to attend to \emph{both} image and other text tokens in the same MHA operation.
In addition, both text and image tokens receive  temporal positional embeddings (\eg, RoPE). 
In contrast, in CA layers text tokens only attend to image tokens, and there is traditionally no information on the temporal position of  image tokens relative to the text ones. 
To bridge this gap, we introduce \CAtv, where  text (``t'') tokens attend to the last seen visual (``v'') tokens \textit{as well as to preceding text tokens} in the same window:
\begin{align}
    \resizebox{\columnwidth}{!}{$
    \textbf{CA}_{\text{t}+\text{v}}(x_T \mid x_{K < i \leq T}, y) = 
    \text{MHA}(x_T \mid y_{1\dots N}\, x_{K+1\dots T})\,.
    $}
    \label{eq:ca_tv}
\end{align}
In other words \CAtv can be thought as the SA operation of  \eqref{eq:token_insertion} operating inside local windows, rather than across the whole image-text interleaved sequence.

\myparagraph{(D3) Additional layers.} 
Cross-attention operates as an additive residual update to the self-attention output within each transformer block, effectively doubling the number of attention layers. A natural way to reduce this overhead is to instead replace self-attention layers with cross-attention layers in a subset of transformer blocks. We denote this variant by \CAl. In our experiments, we replace every second SA layer with a CA layer, yielding efficiency gains with only a minor performance drop compared to \CA. Note that not every layer can be replaced by CA, as no text-to-text attention would remain. 
In contrast, \CAtv   layers also computes text-to-text attention, albeit in local windows, and thus constitutes an interesting fused hybrid between CA and SA.

\myparagraph{(D4) Image token updates.}
With token insertion, the image embeddings are updated through the network in the same way as text tokens:  
In every transformer block, they pass through a \textit{feed forward network} (\textbf{D4.1}) and attend to preceding tokens via \textit{self-attention} as in \hcref{eq:token_insertion} (\textbf{D4.2}). 
In contrast, in cross-attention models image embeddings do not receive any persistent updates, and only undergo different KV projections in every cross-attention layer.
This lack of iterative updates may limit the model's ability to further refine image representations and has motivated prior work to enhance CA-based fusion with image updates through dedicated FFNs~\citep{liu2024streamchat}. 
Similarly, we denote by CA+FFNs a variant of CA where image embeddings are updated through the FFNs of the network.
However, as shown in \hCref{tab:ca_vs_ins}, this incurs significant memory costs at training and may require sacrifices to fit the memory constraints (\eg using shorter training sequence lengths or lower image resolution).

\myparagraph{(D5) Multi-images History.} As mentioned in \hCref{sec:preliminaries}, CA operates in local windows such that a text token may only attend to the last seen image.
In contrast, text tokens in insertion-based models can attend to the whole history, including all past images. 
This is a key trade-off of CA: Not retaining past images in the KV cache may be detrimental to the model when dealing with multiple images (\eg videos), but also enables lower memory usage at inference, as we will demonstrate in \hCref{sec:exp-streaming}.
While prior work has considered integrating multiple past image tokens as keys-values in CA layers~\cite{ye2025mplugowl3}, we do not study this setting in this work,  to retain the focus on efficient VLMs. 
Nevertheless, {this lack of explicit visual history is not necessarily a deal-breaker:} Recent work on context compression~\citep{gist,voco,xu2025streamingvlm} suggests an elegant approach to tackle the lack of explicit image history in CA by adding special ``gist tokens'' in the text stream immediately after every image. In CA layers, gist tokens only attend to the image tokens, while in $\text{SA}_\text{text}$ they naturally interact with other text tokens in a standard causal manner. 
Simply put, gist tokens do not have any initial textual semantic meaning and are placed in the text stream as a way to retain a compressed representation of the image they directly follow.
We implicitly rely on this mechanism in all our video experiments: Each window only contains the latest video frame, and a certain number of gist tokens is placed after each frame. In practice, {we directly use the post-image delimiter, often present in chat templates of recent VLMs such as Qwen-VL, to serve this purpose}.
Thus, when generating an answer in a classical video QA setting, the model only has access to the last frame of the video, and all past gist tokens of previous windows. 
Interestingly, we find that even with this compressed visual history, the CA-based VLMs  achieve surprisingly strong performance on video QA tasks.

\myparagraph{Take-aways}. The above discussion reveals that the transition from CA to token insertion can be decomposed into five key design choices (D1-D5). Notably, D1-D3 can be addressed with simple, lightweight modifications to the cross-attention mechanism that can even \textit{reduce} computational overhead. In contrast, D4 (updating image tokens through FFNs) incurs significant memory and compute costs during training, and avoiding D5 (retaining all past images in the KV cache) is a critical design choice for streaming applications, where the accumulation of image tokens introduces substantial memory and latency overhead. This is reflected in the training and inference costs reported in \hCref{tab:ca_vs_ins}.

\subsection{Scalable Cross-attention Training}
\label{sec:efficient-impl}

We train our models on a mixture of recent public vision-language datasets, primarily composed of question-answer pairs over single images.
Due to the high disparity in sequence lengths between samples, we employ multimodal sequence packing, as commonly done in LLM training~\cite{yang2025qwen3, dubey2024llama3} and modern VLM pipelines~\cite{bai2025qwen25vl, zhang2025videollama3}. 

To match the desired cross-attention inference behavior, where text tokens only attend to the image of their corresponding window, we employ the block-wise attention implementation of FlashAttention-2~\cite{dao2023flashattention2} in the cross-attention layers during training. 
As illustrated in \hyperref[fig:attention_pattern]{\hCref{fig:attention_pattern}}, we define the attention blocks with the image insertion points acting as natural window delimiters.
Importantly for \CAtv, where text tokens appear as both queries and key-values,  placing the image anywhere other than at the beginning of a window  would break  causality between the text tokens during training as the attention mask is bottom-right aligned in the implementation of FlashAttention-2. 

%% file: figs/attention_pattern.tex
\begin{figure*}[t]
    \centering
    \begin{subfigure}{.46\linewidth}
    \centering
    \subcaptionbox{Insertion}[.48\linewidth]{
    \includegraphics[width=\linewidth]{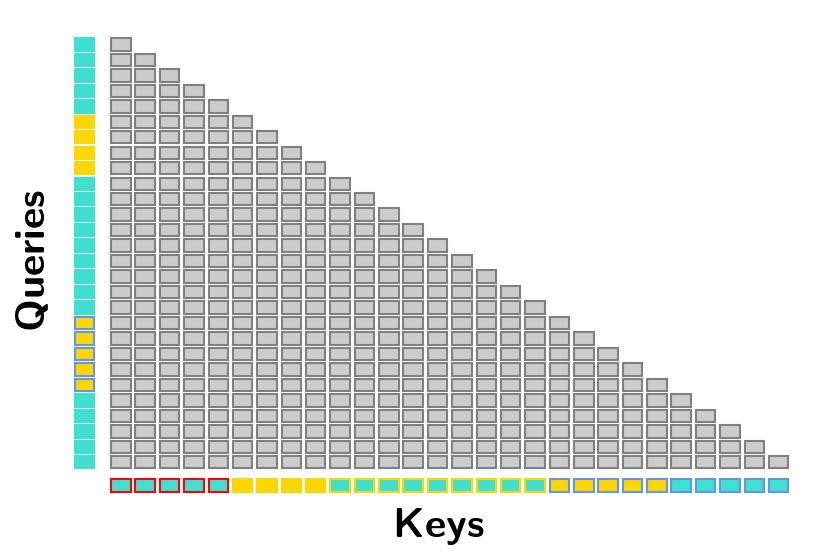}
    }
    \subcaptionbox{Cross-attention}[.48\linewidth]{
    \includegraphics[width=\linewidth]{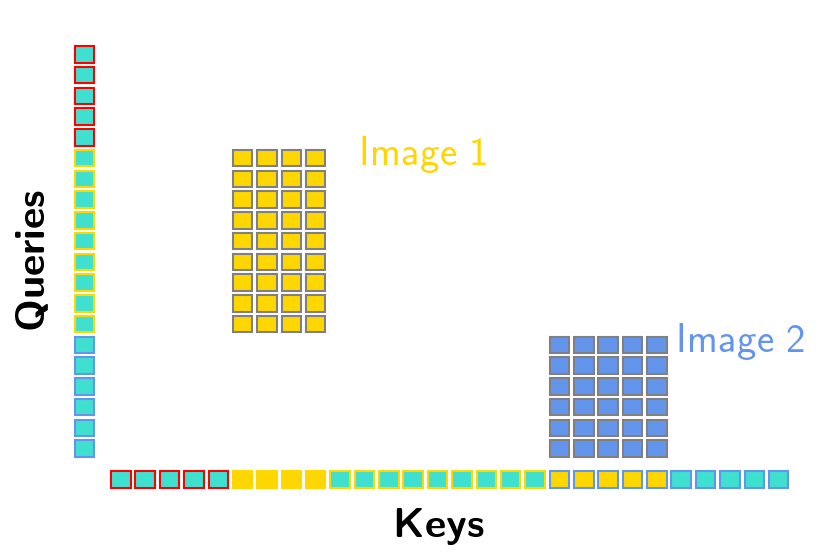} \\
    }
    \subcaptionbox{Text-only SA}[.48\linewidth]{
    \includegraphics[width=\linewidth]{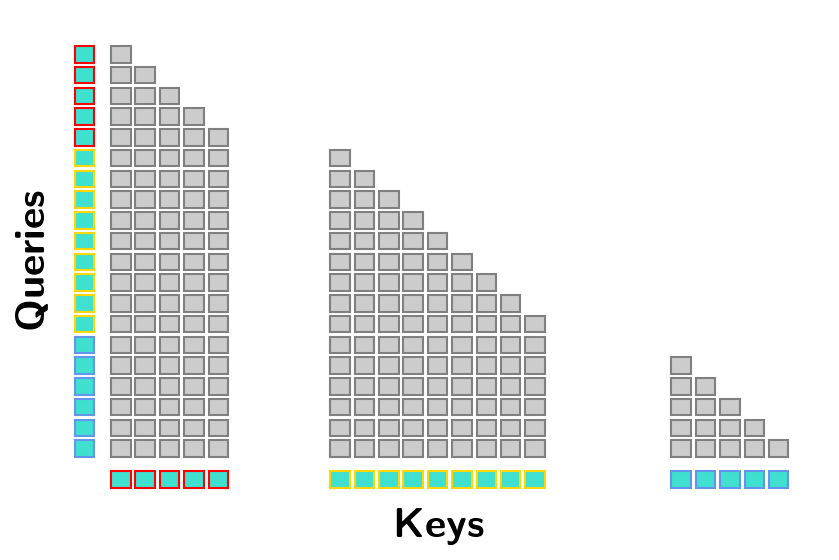}
    }
    \subcaptionbox{\CAtv}[.48\linewidth]{
    \includegraphics[width=\linewidth]{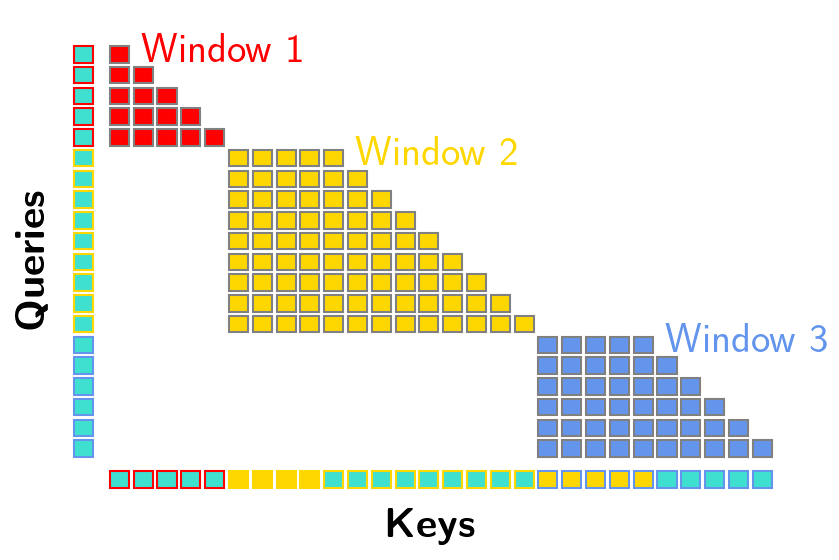}
    }
    \end{subfigure} \hfill
    \begin{subfigure}{.48\linewidth}
    \vspace{.2cm}
    \includegraphics[width=\linewidth]{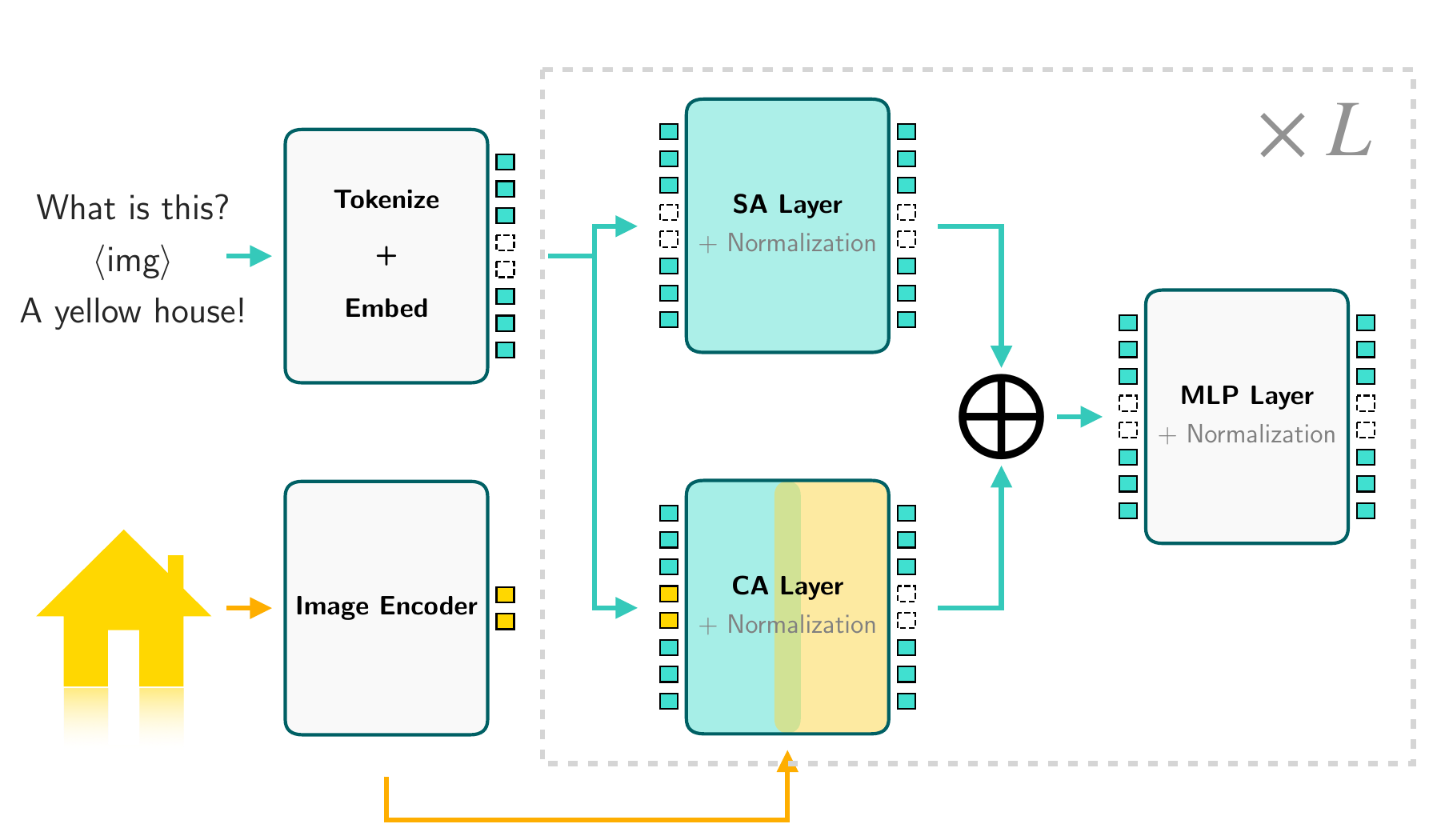}
    \subcaption{Cross-attention model 
    }
    \label{subfig:arch}
    \end{subfigure}
    \caption{
    \textbf{Different attention patterns:} \textbf{(a)} full causal self-attention (SA) on image and text, used in standard insertion-based models, \textbf{(b)} block-wise cross-attention (CA), where each text query attends to all image tokens in its window,  \textbf{(c)} text-only causal SA, used alongside CA to allow for text-to-text interaction, and
     \textbf{(d)} \CAtv, where each text token attends to all preceding text and image tokens in its window (\cf \cref{sec:preliminaries}).
     In \textbf{(e)} we show the general architecture of the CA VLMs we investigate in this work.
    }
    \label{fig:attention_pattern}
\end{figure*}

%% file: tables/table1.tex
\begin{table*}[t]
\centering
\setlength{\tabcolsep}{5pt}
\begin{tabular}{llccccccl}
\toprule
\multicolumn{2}{l}{\multirow{2}{*}{\textbf{Helium1-2B backbone}}} & \multirow{2}{*}{\makecell[c]{Add\\[-2pt]new\\[-2pt]params}} & \multirow{2}{*}{\makecell[c]{Update\\[-2pt]image\\[-2pt]tokens}} & \multicolumn{2}{c}{Training} & \multicolumn{3}{c}{Streaming Inference} \\
\cmidrule(lr){5-6} \cmidrule(lr){7-9}
 & & & & $\text{toks}_{\text{txt}}$/s & Mem. & FPS & Mem. & \;\#KV \\
\midrule
CA                & {\color{gray}{}}                  & {\color{red!70!black}\cmark} & {\color{green!50!black}\xmark} & 1817 & 32.9 & 6.8 & \phantom{0}6.1 & $kT + N$     \\
\CAs              & {\color{gray}{CA + (D1)}}         & {\color{green!50!black}\xmark} & {\color{green!50!black}\xmark} & 1845 & 32.0 & 7.7 & \phantom{0}5.7 & $kT + N$     \\
\CAtv             & {\color{gray}{CA + (D2)}}         & {\color{red!70!black}\cmark} & {\color{green!50!black}\xmark} & 1560 & 33.1 & 5.6 & \phantom{0}6.1 & $kT + (T\!+\!N)$ \\
\CAl             & {\color{gray}{CA + (D3)}}         & {\color{red!70!black}\xmark} & {\color{green!50!black}\xmark} & 1995 & 29.0 & 7.6 & \phantom{0}5.7 & $kT$ or $N$ \\
CA+FFNs        & {\color{gray}{CA + (D4.1)}}       & {\color{red!70!black}\cmark} & {\color{red!70!black}\cmark} & 1504 & 57.8 & 6.6 & \phantom{0}6.1 & $kT + kN$    \\
\midrule
$\text{SA}_{\text{ins}}$ & {\color{gray}{CA + (D1-5)}} & {\color{green!50!black}\xmark} & {\color{red!70!black}\cmark} & 1501 & 62.8 & 1.2 & 29.5 & $kT + kN$    \\
\bottomrule
\end{tabular}
\caption{
\textbf{Efficiency of cross-attention \emph{vs.} token insertion.} We compare cross-attention (CA) and variants with \textbf{(D1)}: shared parameters between CA and SA, \textbf{(D2)}: text-to-text attention in CA layers, \textbf{(D3)}: replacing SA layers with CA, and \textbf{(D4.1)}: image updates through FFNs, and token insertion ($\text{SA}_{\text{ins}}$).
In a  single-image setting, $\text{SA}_{\text{ins}}$ is recovered by adding (D1--5) components to CA.
In all settings, we report whether new parameters are required, whether images embeddings evolve throughout the network, as well as memory and throughput measurements for inference and training.
Measurements are performed on a single GPU. For  inference, we imitate the streaming video captioning setup of \hCref{sec:exp-streaming} in a controlled setting. Specifically, 
we report for how many frames per second (FPS) the resulting model would be able to predict 5 text tokens each, while incorporating the new image information in a streaming fashion.
}
\label{tab:ca_vs_ins}
\end{table*}

%% file: sec/4_experiments.tex
\section{Experiments}
\label{sec:experiments}

\subsection{Experimental Setting}
Below we provide a brief summary of our experimental setup. 
Please refer to \hyperref[app:trainingdetails]{Appendix \ref{app:trainingdetails}} for full details.

\myparagraph{Training Data.} 
We train our models on FineVision ~\citep{wiedmann2025finevision} and a subset of  LLaVA-OneVision-1.5~\citep{an2025llavaov15}. Both  are curated collections of publicly available image-text datasets covering a wide range of tasks such as captioning, document and chart reading, general VQA, etc.
For video training, we further train our models with the aforementioned image-text data alongside LLaVA-Video-178K~\citep{zhang2024llavavid178k}.

\myparagraph{Backbones.} We investigate both extending language-only models with visual understanding as well as adapting existing VLMs with CA. Specifically, we train our models in two settings:
\textbf{(i)} Starting from Helium1-2B~\cite{helium}, a text-only LLM, we jointly finetune  the backbone and \ours layers;
\textbf{(ii)} We adapt a frozen Qwen2.5-VL-3B~\citep{bai2025qwen25vl} VLM by only learning  the additional CA layers. 
In both scenarios, we use the vision encoder of Qwen2.5-VL~\citep{bai2025qwen25vl}. We finetune its last 4 layers when training on image, and freeze it when further finetuning on videos. Finally, we initialize \ours layers from the self-attention layers of the respective backbone. 

\myparagraph{Benchmarks.} 
We evaluate our models on common VLM benchmarks on a variety of tasks: 
document (DocVQA~\citep{mathew2021docvqa}) and  chart (ChartQA~\citep{masry2022chartqa}, InfoVQA~\citep{mathew2022infovqa}) understanding, text recognition (TextVQA~\citep{singh2019textvqa}, OCRBench~\citep{liu2024ocrbench}), and general  QA (RealWorldQA~\citep{realworldqa}, AI2D~\citep{kembhavi2016ai2d}, GQA~\citep{hudson2019gqa}, MME~\citep{Fu2023MMEAC}).
Similarly, we evaluate  video models on video understanding (MVBench~\citep{li2024mvbench}, VideoMME~\citep{fu2025videomme}, PerceptionTest~\citep{patraucean2023perception}), temporal action reasoning (NExT-QA~\citep{xiao2021nextqa}), and long video understanding (MLVU~\citep{zhou2024mlvu}).

\myparagraph{Training Compute.}
As detailed in \hcref{sec:efficient-impl}, we use multimodal sequence packing with block-wise attention during image training, limiting the length of the resulting interleaved text-image sequences to 2048 text tokens and 20,480 image tokens per GPU. We train our models on 64 H100 GPUs with a batch size of 64 and 2 gradient accumulation steps.
For token-insertion experiments, we halve the sequence lengths and double the batch size to fit the increased memory cost while processing the same number of tokens as CA models.
We process images at 
native resolution up to $952^2$ 
pixels (1156 tokens per image) and downscale larger images, keeping their aspect ratio, to this target resolution. 
For videos, we use $R_\text{max}\myeq 504^2$ and extract 2 frames per second, with a maximum clip duration of 3 minutes, resulting in up to 46,080 image tokens per sequence. 
In terms of training horizon, we train for 40k training steps for Helium1 experiments, 25k 
for Qwen2.5-VL adaptation on images, and a further 15k for videos.

\subsection{From LLM to VLM with Cross-Attention}
\label{sec:exp-training}
\input{tables/helium_images}

We first train the text-only Helium1-2B with additional \ours layers. 
In \hCref{tab:sota-images} we compare a Helium1-based VLM trained with token insertion against different CA variants, all trained under the same conditions. 
For reference, we also report performance of  proprietary VLMs  (InternVL2.5~\citep{chen2024internvl25}, Qwen2-VL~\citep{wang2024qwen2vl}, and Video-LLaMA3~\citep{zhang2025videollama3}), an open-source VLM trained on public data  (SmolVLM~\citep{marafioti2025smolvlm}) and recent CA-based VLMs (mPLUG-Owl3~\citep{ye2025mplugowl3}, StreamChat~\citep{liu2024streamchat}). 

As shown in \hCref{tab:sota-images}, the performance of the vanilla cross-attention model, as well as its variants, is very close to the token insertion model trained in the same setting, with an average 1.5 percent drop of performance. 
A significant gap only remains on ChartQA and InfographicVQA, both dealing with understanding complex graphics and figures.
Compared to the literature, our CA models outperform the cross-attention-based mPLUG-Owl3 on most benchmarks, even surpassing its 7B variant, highlighting the importance of an up-to-date training  pipeline to fairly compare CA models with token insertion. 
When comparing different variants of cross-attention, we first note that adding text tokens to the key-values of cross-attention (\CAtv) performs similarly to the vanilla CA on average, with a slight boost on general VQA benchmarks. 
Second, we observe that sharing CA and SA parameters (\CAs) is detrimental to the model performance, but only by a small margin. This introduces an interesting efficiency-performance trade-off, as \CAs is more practical than \CA, according to the results of \hCref{tab:ca_vs_ins}. 
Finally, replacing every other self-attention layer with CA layers (\CAl) provides another solid option to trim off compute while incurring a small drop in performance.
The robustness of CA across these variants is encouraging from a practical standpoint: the most efficient ones (\CAs, \CAl) process over 6$\times$ more frames per second and use over 5$\times$ less memory at inference than token insertion (\hcref{tab:ca_vs_ins}), at only a minor performance cost. We further showcase these advantages on the task of live video captioning in \hcref{sec:exp-streaming}.

\input{tables/qwen_images}
\input{tables/qwen_videos}

\subsection{Adapting an Insertion VLM to Cross-Attention}
\label{sec:exp-adaptation}

The previous results show that CA is competitive with token insertion when both are trained from scratch. In practice, however, adapting strong pretrained VLMs is more efficient.
We therefore investigate whether a pretrained insertion-based model can be efficiently converted to cross-attention. Specifically, we adapt Qwen2.5-VL-3B~\citep{bai2025qwen25vl} by replacing its token-insertion mechanism with CA layers.
We train the added CA layers alongside the four last blocks of the visual encoder and keep all other parameters frozen.

\myparagraph{Image evaluation.} 
We report results in \hCref{tab:qwen-images}, directly comparing to the base model Qwen2.5-VL. 
As in \hCref{sec:exp-training}, changing to a CA-based design incurs a moderate performance drop (6.8\% on average).
Consistent with the trends observed in the previous section, the most significant gap occurs on InfographicVQA.
Importantly, these results are obtained by training only the CA layers and the last image encoder blocks for 25k steps, while keeping all other parameters frozen. Despite this lightweight adaptation, the resulting CA model retains the vast majority of the base model's capabilities while gaining the practical efficiency advantages of cross-attention at inference time.

\myparagraph{Video evaluation.} 
We further finetune our Qwen-CA model on videos~\cite{zhang2024llavavid178k} and report results in \hCref{tab:qwen-video}. 
As discussed in \hCref{sec:differences} (D5), we implement cross-attention with windows, such that each text token only attends to the most recent video frame. 
Thus, to preserve information from past images in the text stream, we rely on the post-image delimiter tokens, already inserted by Qwen-VL's chat template after each frame, to act as gist tokens. 
In particular, this differs from past CA works, which generally insert multiple past frames as key-value inputs in cross-attention layers, thus increasing the cost of the CA operation. 
Despite this  limited visual history, our CA model lands only 3.9 percent below the base model on average, suggesting that it effectively stores relevant visual information in the gist tokens.
For reference, we also train a model {in the standard evaluation setting,} where all image frames are inserted in a window (second-to-last row). 
This leads to a performance boost (but a higher compute cost) and performs on-par or better than prior CA works of larger size. 

\subsection{Ablation Experiments}
\label{sec:exp-ablation}
For ease of reading, 
we group the 9 benchmarks into 3 categories in our ablation results: \textsc{hres}, high-res.\ chart and document reading (DocVQA, InfoVQA, ChartQA), \textsc{ocr}, reading text in natural images (OCRBench, TextVQA), and  \textsc{vqa}, general-knowledge visual understanding (RealWorldQA, AI2D, GQA, MME).

\input{tables/ablation_fusion}

\myparagraph{Cross-attention layer frequency.}
In \hyperref[tab:ablation-fusion]{\Cref{tab:ablation-fusion} (left)}, we report results for training \CAl from the text-only Helium1-2B model~\citep{helium}  where CA layers \emph{replace} SA every \emph{n} layers, with $n = 2$ and $4$. 
Note that the \CAl results of \hCref{tab:sota-images} use $n=2$ by default.
Reducing the number of CA layers 
(here, replacing SA every 4 layers instead of every 2) further shifts the efficiency–performance trade-off, 
providing a simple way to adapt to different compute budgets.

\myparagraph{Updating image tokens.}
\input{tables/ablation_mlp}
Recent CA-based VLMs \cite{liu2024streamchat} update the image embeddings by applying dedicated FFN layers to improve performance.
We evaluate the benefits of this approach through a small-scale ablation with the image encoder frozen, as propagating the image tokens through FFNs substantially increases memory usage  (\cf \hCref{tab:ca_vs_ins}).
As shown in \hyperref[tab:ablation-mlp]{Table~\ref{tab:ablation-mlp}}, updating image tokens yields improves performance ($\approx$\!\,\,2 percents on the average performance), in line with prior work~\citep{liu2024streamchat}, but comes with significantly higher memory usage during training (\hCref{tab:ca_vs_ins}).

\myparagraph{Link to token compression.} 
To reduce token insertion costs, it is common to compress the number of image tokens before inserting them in the text stream~\citep{li2023blip2, zhang2023videollama, maaz2023videochatgpt, li2024videochatflash, wang2025internvideo2}. 
In \hyperref[tab:ablation-fusion]{\Cref{tab:ablation-fusion} (right)}, we report results of training Helium1-2B with either full token insertion or a Q-Former-based compression~\cite{li2023blip2}, in which a small transformer block is applied to the $N$ image tokens produced by the vision encoder, alongside $Q \ll N$ learnable queries; only the $Q$ queries are then inserted in the LLM's textual stream.
For general VQA tasks, we find that even aggressive token compression has limited impact on the performance, but for tasks requiring more detailed representations (\textsc{hres}, \textsc{ocr}) we observe significant performance drops.
Hence, compressed insertion can be a practical alternative to full token insertion if the task does not involve fine-grained visual detail. 
Nonetheless, as we discuss in \hyperref[app:live-captioning]{Appendix \ref{app:live-captioning}}, even with token compression the  memory cost and context length limit quickly become a bottleneck when dealing with long streaming video understanding.
Finally, note that token compression is orthogonal and can be combined with CA to further reduce the number of tokens in the CA layers. 

\subsection{Application to Live Video Captioning}
\input{figs/streaming}
\input{figs/timings}
\label{sec:exp-streaming}
\input{sec/livecc}

%% file: tables/helium_images.tex
\begin{table*}[t]
\centering
\scalebox{1.}{
\setlength{\tabcolsep}{2pt}
\begin{tabular}{lcccccccccc}
\toprule
& \# train & \multicolumn{3}{c}{High-res Doc/Chart} & \multicolumn{2}{c}{Scene Text} & \multicolumn{4}{c}{Knowledge / General QA} \\
\cmidrule(lr){3-5} \cmidrule(lr){6-7} \cmidrule(lr){8-11} 
& tokens & \chartqa & \docvqa & \infovqa & \ocrbench & \textvqa & \realworldqa & \aitd & \gqa & \mme \\
\midrule
\multicolumn{4}{l}{\textit{Token Insertion -- Proprietary}} \\
\midrule
InternVL2.5~\citep{chen2024internvl25} & 0.5T & 79.2 & 88.7 & 60.9 & 804  & 74.3 & 60.1 & 74.9 & 59.5 & 2005 \\
Qwen2-VL~\citep{wang2024qwen2vl} & 1.4T  & 73.5 & 90.1 & 65.5 & 767 & 79.7 & 62.9 & 69.9 & 59.8 & 1872 \\
VideoLLaMA3~\citep{zhang2025videollama3} &    & 79.8 & 91.9 & 69.4 & 779  & \rep{80.1} & 67.3 & 78.2 & 62.7 & 1901 \\
\midrule
\multicolumn{4}{l}{\textit{Token Insertion -- Public data}} \\
\midrule
SmolVLM~\citep{marafioti2025smolvlm} &   & 68.7 & 80.0 & \rep{42.2}   & 729 & 73.0 & \rep{51.0} & \rep{59.7} & \rep{49.2} & \rep{1568} \\
\textbf{Insertion$_\text{He-2B}$ (ours)} & 0.13T & 80.0 & 87.2 & 55.3 & 745 & 74.4 & 59.6  & 64.6 & 54.0  &  1676 \\
\midrule
\multicolumn{4}{l}{\textit{Cross-attention SotA -- Public data}} \\
\midrule
mPLUG-Owl3 8B~\citep{ye2025mplugowl3} & 0.1T & \rep{59.2}& \rep{55.9} & \rep{36.8}&     \rep{527} & 69.0 & \rep{63.9} & 73.4 & 65.0 & \rep{1940}\\
StreamChat 7B ~\citep{liu2024streamchat} &   & $\diamond$ & $\diamond$ & $\diamond$ & $\diamond$ & 72.4 & 61.7 & 76.6 & 62.4 & 1520 \\ 
mPLUG-Owl3 2B & 0.1T  & \rep{48.5}& \rep{48.2} & \rep{28.1}& \rep{450}& 62.6 & \rep{56.9} & 62.6 & 61.0 & \rep{1551} \\
\midrule
\multicolumn{4}{l}{\textit{Cross-attention (\textbf{ours}) -- Public data}} \\
\midrule
\textbf{\CAHe} & 0.13T & 75.9  &  85.8  &  51.8  &  722  &  73.8  &  58.0  &  66.2  &  54.3  &  1731  \\
\textbf{\CAsHe} & 0.13T & 74.1 & 84.0 & 50.6 & 696 & 73.4 & 56.5 & 64.2 & 51.7 & 1613 \\
\textbf{\CAlHe} & 0.13T & 73.8  &  84.5  & 48.0  &  714  &  73.0 &  56.9  &  64.3 &  54.0  &  1607\\
\textbf{\CAtvHe} & 0.13T & 75.3& 85.1& 52.1& 722& 74.1 & 60.3 & 65.5 & 54.5 & 1720 \\
\bottomrule\\[-.8em]
\multicolumn{6}{l}{\rep{}: \footnotesize  Reproduced with the publicly available models at \href{https://huggingface.co}{huggingface}.} & \multicolumn{5}{l}{$\diamond$: \footnotesize  Results 
not 
available.} 
\end{tabular}
}
\caption{
\textbf{Comparing cross-attention and insertion-based models}. Unless specified, all  models are built on \textbf{2B} LLMs. We use \texttt{lmms-eval}~\cite{zhang2025lmms} for evaluation, and re-evaluate existing models when benchmark results are not provided in the original work. 
When trained under the same conditions, our CA model reaches 
{performance close to that of} a token insertion model, showing the potential of cross-attention {as an efficient alternative}. Nonetheless, a significant gap remains on infographic and chart understanding tasks, highlighting the 
benefits of token insertion for certain tasks. 
\vspace{-.1cm}
}
\label{tab:sota-images}
\end{table*}

%% file: tables/qwen_images.tex
\begin{table*}[t]
\centering
\scalebox{0.9}{
\setlength{\tabcolsep}{3pt}
\begin{tabular}{lccccccccccc}
\toprule
& \multicolumn{3}{c}{Document/Chart} & \multicolumn{2}{c}{Scene Text}  & \multicolumn{4}{c}{Knowledge / General QA} \\
\cmidrule(lr){2-4} \cmidrule(lr){5-6} \cmidrule(lr){7-10}
 & \chartqa & \docvqa & \infovqa & \ocrbench & \textvqa & \realworldqa & \aitd & \gqa & \mme \\
\midrule
\multirow{2}{*}{Qwen2.5-VL 3B} & 84.0 & 93.0 & 77.1 & 797 & 79.3 & 65.4 & 81.6 & $\diamond$ & 2157 \\
 & \rep{83.1} & \rep{92.4} & \rep{75.1} & \rep{796} & \rep{79.6} & \rep{60.4} & \rep{79.6} & \rep{61.0} & \rep{2224} \\
\midrule
\textbf{\CAQw{}} & 80.3 &  87.0  &  57.4  &  783  &  76.3  &  61.3 &  74.0  &  59.0  &  1910 \\
\textbf{\CAtvQw{}} & 81.3 & 87.1 & 56.8 & 775 & 76.3 & 63.8 & 74.4 & 58.6 & 1971 \\
\bottomrule \\[-.8em]
\multicolumn{6}{l}{\rep{}: \footnotesize  Reproduced with the publicly available model at \href{https://huggingface.co}{huggingface}.}
\end{tabular}
}
\vspace{-.1cm}
\caption{\textbf{Adapting a frozen Qwen2.5-VL} by training additional CA layers (and the last 4 block of the image encoder) retains similar  performance as  the base model on most benchmarks. As for the experiment of \hCref{tab:sota-images}, a large gap remains on the InfographicVQA benchmark, suggesting some tasks may benefitting from token insertion fusion mechanism than others. }
\vspace{-.2cm}
\label{tab:qwen-images}
\end{table*}

%% file: tables/qwen_videos.tex
\begin{table*}[t]
\centering
\scalebox{1.}{
\begin{tabular}{lcccccccc}
\toprule
\small 
& \sc Max & \multirow{2}{*}{\sc Windows} & \multicolumn{2}{c}{\sc VideoMME} & {\sc {Next}} & {\sc {Percep.}} & {\sc {MV}} & \multirow{2}{*}{\sc {MLVU}} 
\\
 & \sc{\#Frames} & & w/o sub & w/ sub & \sc{QA} & \sc{Test} & \sc{Bench} & 
 \\
\midrule
\multirow{2}{*}{Qwen2.5VL 3B} & 768 & \multirow{2}{*}{\xmark} &  61.5 & 67.6 & $\diamond$ & 66.9 & 67.0 & 68.2 
\\
 & 360 &  &  \rep{58.8} & \rep{63.6} & \rep{78.9} & \rep{67.0} & \rep{66.2} & \rep{65.2} 
 \\
\midrule
StreamChat 7B \; & 40 & \xmark & 58.6 & 62.8 &  78.5 & $\diamond$ & 53.3 & 63.9 \\
mPlug-Owl3 7B & 128 & \xmark & 53.5 & $\diamond$ & 78.6 & $\diamond$ & 54.5 & $\diamond$ \\
mPlug-Owl3 7B$^\dagger$ & 180 & \xmark & \rep{59.8} & \rep{65.2} & \rep{82.1} & \rep{65.6} & \rep{58.3}  & \rep{65.6} \\
\CAtvQw{}  3B & 180 & \xmark & 59.0 & 64.8 & 79.8 & 63.8 & 60.6 & 67.9 \\
\midrule
\CAtvQw{} 3B & 180 & \cmark & 56.9 & 60.5 & 78.8 & 62.6 & 59.5 & 66.2 \\
\bottomrule \\[-.8em]
\multicolumn{7}{l}{\rep{}: \footnotesize Reproduced with the publicly available model at \href{https://huggingface.co}{huggingface}.}
\end{tabular}
}
\vspace{-.1cm}
\caption{\textbf{Video benchmarks.}
We further finetune our Qwen2.5-VL model with CA layers on the LlavaVid dataset~\cite{an2025llavaov15} for 15k training steps.
Results are reported for \CAtv in a setting where each window contains a single frame (\cmark), using gist tokens to preserve past visual history, as described in \hCref{sec:differences}. 
For fair comparison, we also report results in the more standard, yet less efficient, evaluation setting where all frames are placed in a single window (\xmark), as done in prior CA work. In both scenarios, adapting Qwen2.5-VL to use cross-attention layers 
only incur a small performance drop and performs on par with prior CA models, even those of larger size.}
\vspace{-.2cm}
\label{tab:qwen-video}
\end{table*}

%% file: tables/ablation_fusion.tex
\begin{table}[t]
\centering
\setlength{\tabcolsep}{4pt}
\footnotesize
\scalebox{.9}{
\begin{tabular}{lcccccc}
\toprule
& \multicolumn{3}{c}{\textit{(a) \CAtvl{} layer {frequency}}} & \multicolumn{3}{c}{\textit{(b) Impact of token compression}} \\
\cmidrule(lr){2-4} \cmidrule(lr){5-7}
& \multirow{2}{*}{CA}
& \multicolumn{2}{c}{\CAl (every $n$ layers)}
& \sc{Insertion}
& \sc{QFormer}
& \sc{QFormer}
\\
\sc{Task} & & $n=2$ & $n=4$ & (1156 tokens) & (128) & (32) \\
\midrule
\sc{hres}
& 70.8 & 68.4 & 67.8
& \textbf{74.2} & 62.7 & 56.0 \\
\sc{ocr}
& 73.1& 72.0 & 71.1
& \textbf{74.5} & 70.8 & 67.0 \\
\sc{vqa}
& 60.4 & 58.8 & 57.0
& \textbf{60.6} & 57.8 & 56.9  \\
\sc{avg.}
& 68.1& 66.4 & 65.3
& \textbf{69.8} & 63.8 & 59.9 \\
\bottomrule
\end{tabular}
}
\caption{
\textbf{Ablation experiments on (a) \CAl and (b) token compression.}
{\textbf{(a)}} {Effect of inserting \CAl at different layer intervals, in the same training setting as} \hyperref[tab:sota-images]{Table \ref{tab:sota-images}.}
{\textbf{(b)}} 
Impact of token compression, a common approach to improve the efficiency of token insertion.
Reducing the number of image tokens quickly deteriorates performance on tasks involving high resolution images. Furthermore, as we show in \hyperref[sec:exp-streaming]{Section \ref{sec:exp-streaming}}, token compression is not sufficient to solve the memory bottleneck inherent to image token insertion, as the KV cache {still} grows during streaming inference.
}
\label{tab:ablation-fusion}
\end{table}

%% file: tables/ablation_mlp.tex
\begin{table}[t]
\centering
\footnotesize
\begin{tabular}{lccccc}
\toprule
    \sc{Model  \textbackslash\  Benchmark } && \sc{hres} & \sc{ocr} & \sc{vqa} & \sc{avg.} \\
    \midrule
    \CAtv (He-2B) &&  55.7  &  64.1  &  \textbf{44.1}  &  54.7  \\  
    \hspace{0.15cm}\sc{ + ffn updates}  &&  \textbf{58.1}  &  \textbf{66.6}  &  43.2  &  \textbf{56.0}  \\  
\bottomrule
\end{tabular}
\caption{\textbf{Updating image embeddings} through the underlying LLM's FFNs provides a modest increase in performance, but incurs non-negligible memory and compute costs. 
We also freeze the image encoder to accommodate the increased memory usage.}
\vspace{-.2cm}
\label{tab:ablation-mlp}
\end{table}

%% file: figs/streaming.tex
\begin{figure*}
    \centering
    \hspace{-.5cm}
    \includegraphics[trim=1.5em 0 0 0,clip,width=\textwidth]{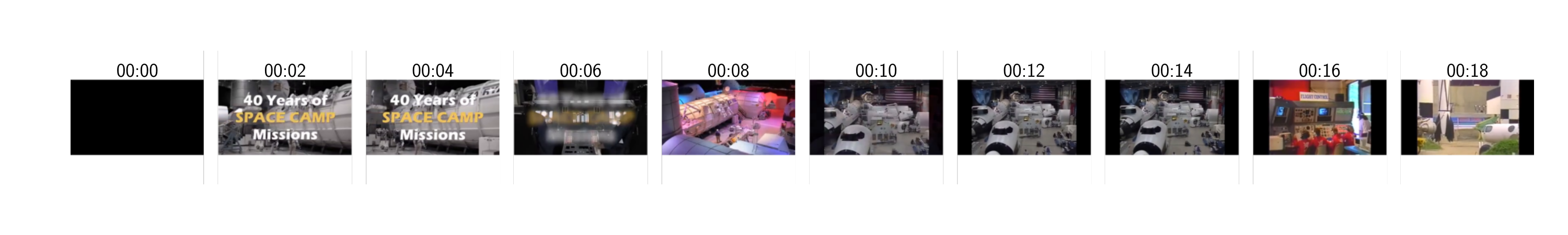}
    \vspace{-.7cm}
    \input{figs/streaming/40_years_of_space_camp_subtitles}
    \vspace{-.2cm}
    \caption{\textbf{Live captioning}. 
    We display captions generated by our \CAtvQw model. 
    Each text span is annotated with the corresponding frame's timestamp (\textit{top})   and the model's delay as  {\timestamp{\color{black}timestamp $|$ \color[rgb]{0, .5, 0}delay}}. 
    As shown in \hyperref[subfig:delays]{Figure \ref{subfig:delays}}, the model's outputs are generated much faster than real-time with no noticeable increase over time and near-constant memory usage, as only new text tokens are added to the model's KV cache; for further qualitative examples, including insertion-based VLMs, see \hyperref[sec:exp-streaming]{Appendix ~\ref{app:live-captioning}}.
    }
    \label{fig:qualitative}
    \vspace{-.5em}
\end{figure*}

%% file: figs/streaming/40_years_of_space_camp_subtitles.tex
\begin{subtitlebox}
\footnotesize
\textit{\footnotesize This video shows}
\timestamp{00:01$|$\color[rgb]{.1, .5, 0}0.0s}~\textit{\footnotesize the Apollo 13}
\timestamp{00:02$|$\color[rgb]{.1, .5, 0}0.0s}~\textit{\footnotesize spacecraft during its}
\timestamp{00:03$|$\color[rgb]{.1, .5, 0}0.0s}~\textit{\footnotesize 40 years of space camp missions at}
\timestamp{00:05$|$\color[rgb]{.1, .5, 0}0.0s}~\textit{\footnotesize NASA's Johnson Space Center}
\timestamp{00:06$|$\color[rgb]{.1, .5, 0}0.0s}~\textit{\footnotesize in Houston Texas}
\timestamp{00:07$|$\color[rgb]{.1, .5, 0}0.0s}~\textit{\footnotesize and it was filmed by NASA}
\timestamp{00:08$|$\color[rgb]{.1, .5, 0}0.0s}~\textit{\footnotesize Space Camp Director}
\timestamp{00:09$|$\color[rgb]{.1, .5, 0}0.0s}~\textit{\footnotesize Steve Smith who has been with us for over}
\timestamp{00:10$|$\color[rgb]{.1, .5, 0}0.0s}~\textit{\footnotesize 25 years now}
\timestamp{00:11$|$\color[rgb]{.1, .5, 0}0.0s}~\textit{\footnotesize at this point he had to go back home}
\timestamp{00:12$|$\color[rgb]{.1, .5, 0}0.0s}~\textit{\footnotesize to his job as a}
\timestamp{00:13$|$\color[rgb]{.1, .5, 0}0.0s}~\textit{\footnotesize director but we're going}
\timestamp{00:15$|$\color[rgb]{.1, .5, 0}0.0s}~\textit{\footnotesize with him here today}
\timestamp{00:16$|$\color[rgb]{.1, .5, 0}0.0s}~\textit{\footnotesize so you can see how they've changed}
\timestamp{00:17$|$\color[rgb]{.1, .5, 0}0.0s}~\textit{\footnotesize over time from}
\timestamp{00:18$|$\color[rgb]{.1, .5, 0}0.0s}~\textit{\footnotesize Apollo era}
\timestamp{00:19$|$\color[rgb]{.1, .5, 0}0.0s}~\textit{\footnotesize missions all right}
\timestamp{00:20$|$\color[rgb]{.1, .5, 0}0.0s}~\textit{\footnotesize we'll start}
\end{subtitlebox}

%% file: figs/timings.tex
\begin{figure*}[t]
    \centering
\begin{subfigure}{0.4\linewidth}
        \centering
        \small 
        \resizebox{\linewidth}{!}{ 
        \begin{tabular}{lcc}
            \toprule
            Model & Training steps & Winrate (\%) \\
            \midrule
            \multicolumn{3}{l}{\textit{\CAtv (Qwen2.5-VL)  (Ours, 3B)}} \\
            \midrule
            & 3k  & 25.4 \\
                                   & 6k  & 27.6 \\
                                   & 8k  & 33.7 \\
                                   & 12k & 34.3 \\
                                   & 15k & 36.3 \\
                                   & 17k & 39.4 \\
                                   & 20k & 39.0 \\
            \midrule
            \multicolumn{3}{l}{\textit{Baselines as reported in LiveCC, $7B$}} \\
            \midrule
            LiveCC-7B-Base     & & 43.2 \\
            LiveCC-7B-Instruct & & 41.5 \\
            Qwen2-VL-7B-LiveCC & & 33.7 \\
            \bottomrule
        \end{tabular}}
        \subcaption{LLM-as-judge evaluation on LiveSports video captioning}
    \end{subfigure}
    \hfill 
    \begin{subfigure}{0.55\linewidth}
        \centering
        \includegraphics[width=\linewidth]{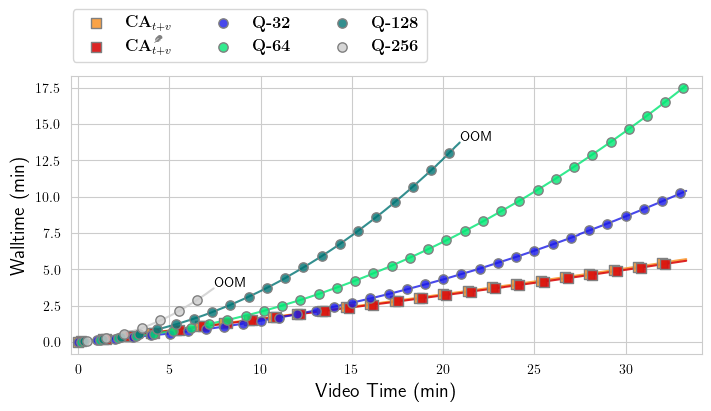}
        \subcaption{Walltimes in the  streaming video captioning setting for CA and token-insertion}
        \label{subfig:delays}
    \end{subfigure}
    \caption{\textbf{Quantitative results on LiveCC.} \textbf{(a)} We evaluate our \CAtvQw model on the LiveSports captioning tasks proposed in \cite{chen2025livecc}, following their LLM-as-a-judge methodology. \textbf{(b)} We record the walltime as a function of the number of frames inserted in a streaming captioning scenario, for \CA and token-compression techniques (Q-Former with different numbers of query tokens). 
    While token compression mitigates the computational cost of token insertion for short videos, the CA model maintains low latency inference for longer times. In addition, token insertion tends to go out-of-memory quickly especially for larger numbers of tokens. Note that for better readability, we only plot a subset of markers, although the  plotted measurements occur at every frame.}
    \vspace{-.1cm}
    \label{fig:dalys}
\end{figure*}

%% file: sec/livecc.tex
Streaming video understanding is an important motivation for building efficient vision-language models.
This tasks comes with two particular challenges: \textbf{(i)} Memory costs needs to be kept under  control  in order to handle longer videos, and \textbf{(ii)} for streaming applications, the model latency needs to stay below the frame rate to avoid accumulating delay over time. 
By design, CA is well equipped to address both of these issues. To illustrate the usefulness of CA to process long text-image sequences,  we now consider the task of live video captioning.

Specifically, we finetune our \CAtv-adapted Qwen2.5-VL model trained on images only on
Live-WhisperX-526K, a recent instruction-tuning dataset~\citep{chen2025livecc}. It is composed of video frames sampled at
2fps and interleaved with the text transcript of the video's original audio. 
As discussed in \hCref{sec:differences}, we design our CA layers such that text tokens only attend to the latest image in the current window. At inference, this means we simply continuously execute the model while replacing the key-value sources of the CA layers every half a second.
We provide qualitative samples of  live captioning results in  \hyperref[fig:qualitative]{Figure \ref{fig:qualitative}} and  \hyperref[app:live-captioning]{Appendix \ref{app:live-captioning}}.

\myparagraph{Quantitative evaluation.} 
We evaluate our LiveCC-tuned model on 
the LiveSports3K benchmark proposed in \cite{chen2025livecc}. The dataset consists of videos of sports events ($\sim$ 20 seconds long).
The captions are evaluated using an LLM as a judge, following the evaluation protocol provided in LiveCC's repository with GPT-4o acting as the judge.
In \hyperref[fig:dalys]{Table \ref{fig:dalys}(a)}, we report results for a \CAtv Qwen2.5-VL-3B model trained on LiveCC and evaluated across training steps, and compare to the results reported in the original LiveCC paper. 
Notably, despite its smaller size (3B \vs 7B), our CA model obtains scores similar to those of LiveCC.

\myparagraph{Real-world performance.} 
As can be seen in \hyperref[fig:dalys]{Figure \ref{fig:dalys}(b)}, the memory cost of token insertion methods increases more rapidly than for the CA-based model.  
While token compression reduces the cost of token insertion for short conversations, it cannot alone prevent the increased memory usage leading to OOM when the number of tokens is too high. 
In \hyperref[fig:dalys]{Figure~\ref{fig:dalys}}, we also report the wall-time of the same models (recorded on a single H100 GPU) as a function of the number of frames inserted: 
Generation with insertion-based models becomes progressively slower as image tokens accumulate in the KV cache.
In contrast, the CA-based model maintains high inference speed over much longer horizons.
Together, these results show that the efficiency properties of CA translate into tangible advantages for real-world streaming, enabling low-latency captioning over extended time horizons where insertion-based models are impractical.

%% file: sec/conclusion.tex
\section{Conclusions}

We revisit CA as a fusion mechanism
for VLMs, a design that has been largely set aside in favor of token insertion despite CA's favorable efficiency properties.
We identify five core design differences (D1-D5) that progressively bridge CA and token insertion, and analyze their respective impact on efficiency and performance.
Training CA-based VLMs both from a text-only LLM and by adapting a pretrained insertion-based model, we find that simple cross-attention performs close to token insertion on most image and video benchmarks, even without employing 
architectural modifications proposed in prior work.
As a result, our experiments on adapting a pretrained SotA  insertion-based VLM to cross-attention yield 
a strong CA-based model, outperforming prior CA-based VLMs of larger size (\cref{fig:fig1}). 
This adapted model naturally lends itself to streaming applications: by replacing the key-value sources of the CA layers at each new frame, it performs live video captioning
with near-constant memory and latency, while token insertion models quickly exhaust their memory budget.
Our results suggest that cross-attention deserves renewed consideration as a practical and competitive alternative for vision-language fusion, especially as applications move toward longer, streaming multimodal inputs.

%% file: sec/X_suppl.tex
\newpage
\appendix

\section{Experimental Details}

\subsection{Training Details}
\label{app:trainingdetails}
\myparagraph{Image data.}
We train our image models on FineVision~\citep{wiedmann2025finevision} and a subset of LLaVA-OneVision-1.5~\citep{an2025llavaov15}. 
FineVision and LLaVA-OneVision-1.5 are curated collections of publicly available image–text datasets with 24M images covering captioning, chart reading, grounding and counting, mathematics, document understanding and OCR, and general VQA.
In FineVision, we replace Doclingmatix~\citep{nassar2025smoldocling} with Docmatix~\citep{laurenccon2024docmatix} and over-sample it by a factor of 6. 

As the two datasets overlap substantially,  we only retain a subset of LLaVA-OneVision-1.5 datasets not already included in FineVision: {
{OmniDocBench~\cite{ouyang2025omnidocbench}, allenai-pixmo-docs~\cite{deitke2025molmo}, amc-aime, aops-forum, arxiv-figs, diagram, GQA~\cite{hudson2019gqa}, infographic-azuregpt4v, invoices, latex-ocr, llava-cot-100k, llava-wild, llrv-gpt4v, olympiads, oroikon-chart-captioning, rootsautomation, sherlock, SVIT~\cite{zhao2023svit}, TinyChart~\cite{zhang2024tinychart}, UReader-chart~\cite{ye2023ureader}, UReader-ocr~\cite{ye2023ureader}, UReader-tr~\cite{ye2023ureader}, viquae, vision-oritented, visual-chat, and WIT~\cite{srinivasan2021wit}}}.

We process images at their native resolution using Qwen 2.5-VL's visual encoder.
During training, we use image resolution up to $952^2$ pixels.  Note that the encoder  applies pixel unshuffling, reducing the number of visual tokens by a factor of 4. For instance, a $952^2$ image yields 1156 image tokens, while a $504^2$ frame yields 324. 

\myparagraph{Video training data.}
LLaVA-Video-178K~\citep{zhang2024llavavid178k} is a video instruction-tuning dataset comprising 1.3M question–answer pairs over 178K videos of up to 3 minutes, annotated with open-ended and multiple-choice questions generated using GPT-4o and human input. 
In practice, we sample frames at 2 fps, except for clips shorter than 10\,s, for which we uniformly sample 20 frames. 
We process frames at lower resolutions than for single images ($504^2$ pixels), to compensate for the higher number of images.

\input{tables/supp_config}

\input{tables/supp_resolution}
\myparagraph{Training schedule.}
Training is performed on 64 NVIDIA H100 GPUs with a batch size of 128 for all \CA models (batch size 64 and 2 gradient accumulation steps). 
For token insertion, the feasible sequence length per device is smaller due to higher memory constraints. We thus trade-off shorter sequence lengths (half shorter) with a higher batch size (4 gradient accumulatio steps insteas of 2) to maintain a total sequence length similar to that of CA models. 
Our CA training starting from Helium1-2B takes 25 hours (40k steps), while our adaptation of Qwen2.5-VL-3B takes 16 hours for the image-only training stage (25k steps), and 26 hours for image-video training (15k steps). Training our Insertion$_\text{He-2B}$ model takes 24 hours (40k steps, 4 accumulation steps).
Each GPU processes either a  sequence of image samples (with multimodal sequence packing, as described in \hcref{sec:efficient-impl} or a single video sample.
More specifically, for each batch, we sample either a packed image sequence consisting of multiple question-answer pairs from the image training data (see above) or a single video sample, at a ratio 3:1 (image:video).
The packed image sequence is limited by a maximum number of text and image tokens, specified in \hyperref[tab:supp-config]{Table~\ref{tab:supp-config}}, which also reports the trained parameters for each training stage.

\myparagraph{Optimization.}
We use a standard cross-entropy-based next-token-prediction loss applied only to the answer tokens of a given question-answer pair.
We use AdamW with a constant learning rate schedule apart from a linear warmup and decay.
The learning rate is set to $10^{-4}$ for training new parameters, and $10^{-5}$ for adapting existing parameters.

\subsection{Architecture Details}

\myparagraph{Multimodal sequence packing.}
To train our token-insertion model with sequence packing, we also employ block-wise attention to guarantee that the self-attention operations are properly masked: Each question-answer sample in the packed sequence attends to itself without carrying over any textual context from preceding samples in the sequence. 
This makes the procedure  equivalent to batched training, while being  more efficient as it avoids padding samples to the maximum sequence length within the batch.

\myparagraph{Attention windows in \CA.} As detailed in \cref{sec:differences} and \hcref{fig:attention_pattern}, the attention operation in \CA layers acts in local attention windows, which are naturally delimited by image occurrences: Each window consists of a single image (or multiple \textit{consecutive} images) followed by the associated text. 
Consequently, \textbf{(i)} during text-image training with packed multimodal sequences, windows in \CA layers consist of question-answer pairs and their associated image(s). \textbf{(ii)} When training on LLaVA-Video-178K~\citep{zhang2024llavavid178k}, we define a separate window for each frame as discussed in \hcref{sec:exp-adaptation}: Simply put, each window contains a single frame and their image delimiter tokens from the Qwen's chat template, acting as gist tokens; except for the last window which contains the question and answer text tokens. In \hCref{tab:qwen-video} we also report results closer to standard CA-based models, where all frames extracted from the video are included in a single window, preserving the complete visual history. 
\textbf{(iii)} When training on LiveCC~\citep{chen2025livecc}, each window consists of a single frame (extracted at 2fps) and the corresponding closed captions for this timestamp, which is typically only a few tokens long. Nevertheless, the global coherence of the entire video script  is preserved through the text-only tokens interactions in the self-attention layers.

\myparagraph{Chat template.} For Qwen2.5-VL, we use the model’s provided chat template, which includes pre- and post-image tokens ({\sc$\langle$vision\_\sc{start}$\rangle$} and {\sc $\langle$vision\_\sc{end}$\rangle$}), user–assistant turn delimiters, and a system prompt; we simply omit the insertion of image-token placeholders when training \CA. For Helium1-2B, we use a minimal template in which user and assistant turns are wrapped with their respective start and end tokens, without any pre- or post-image tokens.

\myparagraph{Image processing.} As we rely on the vision encoder of Qwen2.5-VL for all of our models, we apply the corresponding preprocessing for images (patch size of 14, flattening, \etc). Note that for videos, however, we do not use Qwen2.5-VL’s video-processing strategy of temporally downscaling the video frames by a factor of 2, as we found it detrimental in our experiments. Instead, we process videos frame by frame, in the same way as images.

\myparagraph{RoPE in \CAtv.} We do not use any positional embeddings in standard cross-attention layers. For the \CAtv variant which includes text tokens as key-values, we apply the language backbone's RoPE implementation within \CAtv layers, \ie standard RoPE for Helium1-2B and multi-modal RoPE for Qwen2.5-VL with the temporal position of image tokens frozen (they do not advance across image tokens).

\section{Additional Results}
\label{sec:supp-results}

\subsection{Training Speed}
\begin{figure*}[t]
    \centering    \includegraphics[width=\linewidth]{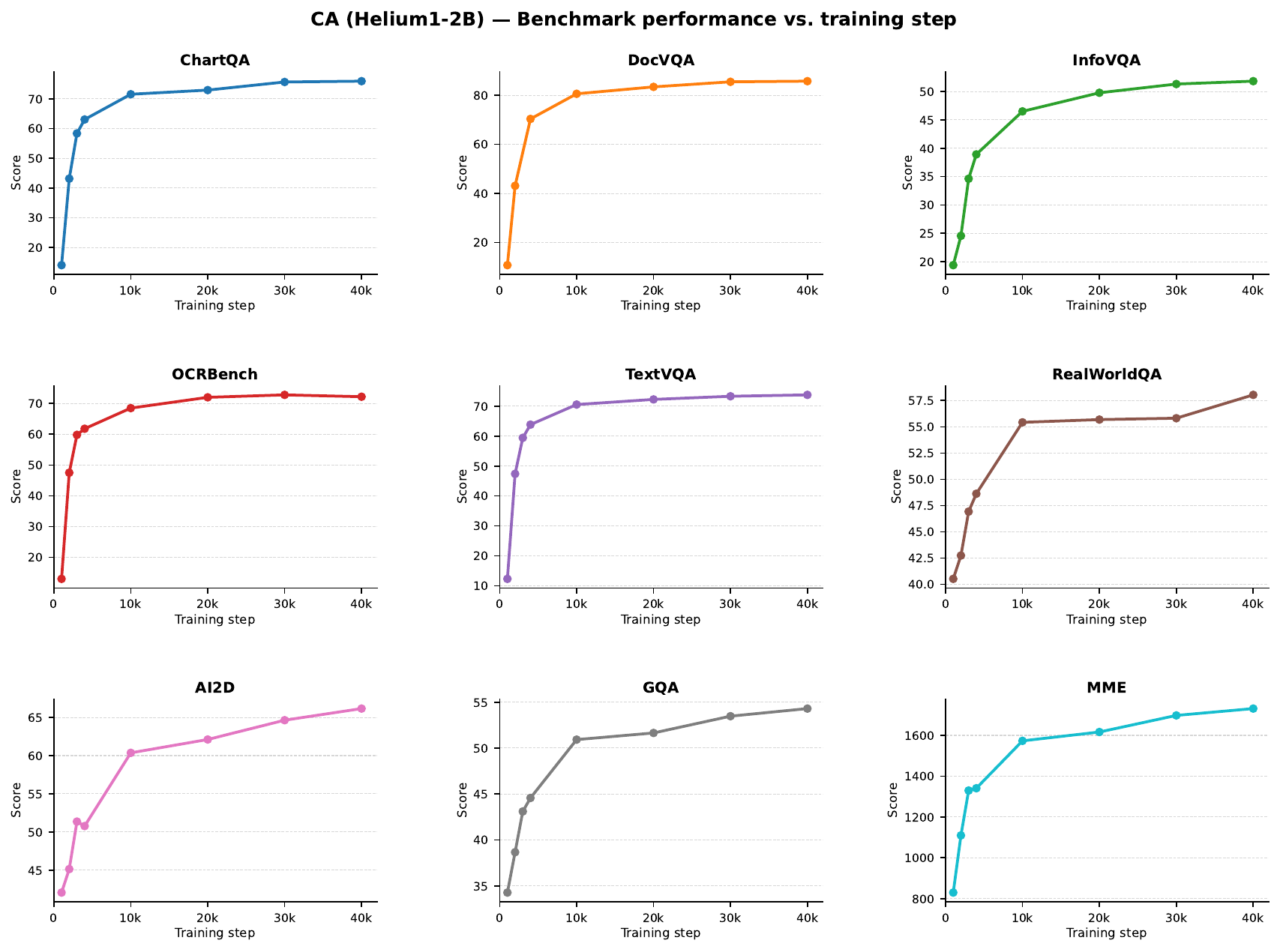} 
    \caption{
    \textbf{Performance across training} for \CA (Helium1 - 2B) across all benchmarks
    }
    \vspace{-.2cm}
    \label{fig:supp_helium_steps}
\end{figure*}

\begin{figure*}[t]
    \centering    \includegraphics[width=\linewidth]{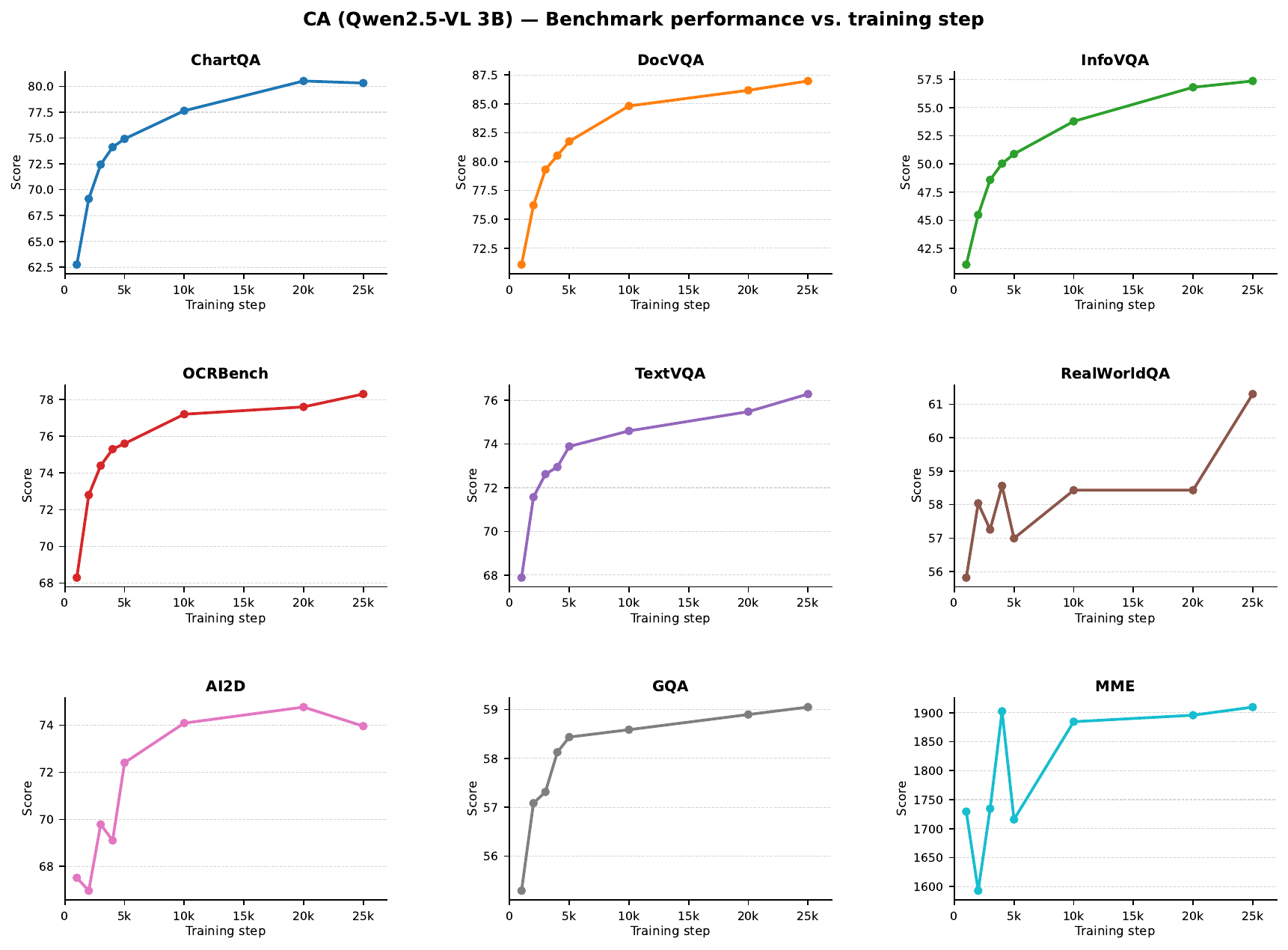} 
    \caption{
    \textbf{Performance across training} for \CA (Qwen2.5-VL 3B) across all benchmarks
    }
    \vspace{-.2cm}
    \label{fig:supp_qwen_steps}
\end{figure*}

In \hCref{fig:supp_helium_steps} and \hCref{fig:supp_qwen_steps}, we report the performance of our \CA models trained from Helium  and adapted from Qwen2.5-VL respectively, at different time during training. 
We find that both settings require few training steps to train the \CA layers, as most performance gains occur within the first 20k training steps for both models. For reference, we report our final results in \hCref{tab:qwen-images} for 25k training steps and in \hCref{tab:sota-images} for 40k training steps.

\subsection{Impact of image resolution}
In \hyperref[tab:sota-images]{Table~\ref{tab:sota-images}}, we report results on VLM benchmarks for \CAHe and Insertion$_\text{He-2B}$ evaluated at maximum resolutions of $952^2$.
In \hCref{tab:supp-resolution}, we additionally show results when varying the maximum resolution at inference ($448^2$, $672^2$, $896^2$, and $1344^2$). 
As expected, resolution can be reduced with little to no loss of performance for general VQA tasks, whereas tasks involving high-resolution images (DocVQA and InfoVQA) exhibit a noticeable performance drop. 

\section{Live Video Captioning}
\label{app:live-captioning}
Finally, we experiment with the task of streaming video captioning to showcase CA's practical benefits. 
To that end, we directly finetune the \CAtvQw model on the recent  Live-Whisper-526K dataset\cite{chen2025livecc} 
for up to 20,000 steps.
We report a set of qualitative videos with caption subtitles generated with \CAtvQw-LiveCC on our \href{kyutai.org/casa}{project page}: The captions are displayed at the time they are generated by the model, to mimic livestreaming conditions. For a static visualization of video excerpts, see \hyperref[fig:qualitative]{Figure \ref{fig:qualitative}} and \hyperref[fig:qualitative2]{Figure \ref{fig:qualitative3}}.

\input{figs/streaming_supp}

%% file: tables/supp_config.tex
\begin{table*}[tbh]
\centering
\footnotesize
\begin{tabular}{l l cc c cc}
\toprule
&  & \multicolumn{2}{c}{Trained params} 
    &  & \multicolumn{2}{c}{Max \#Tokens} \\
\cmidrule(lr){3-4} \cmidrule(lr){6-7}
\multirow{2}{*}{Model} & \multirow{2}{*}{Stage} & Image encoder & \multirow{2}{*}{LLM} & \multirow{2}{*}{\#Steps}  & \multirow{2}{*}{Image} & \multirow{2}{*}{Text} \\ 
 &  & (4 last blocks) & &  &  &  \\ 
\midrule

\multirow{1}{*}{Helium1-2B} 
    & Image stage & \cmark & \cmark & 40k & 20,480 & 2048 \\
\midrule

\multirow{3}{*}{Qwen2.5-VL}
    & Image training      & \cmark  & \xmark & 25k & 20,480 & 2048  \\
    & Image-video training      & \xmark  & \xmark & 15k & 46,080 & 3072 \\
    & LiveCC training     & \xmark   & \xmark & 20k & 30,720 & 3072 \\
\bottomrule
\end{tabular}
\caption{\textbf{Training configurations.} 
Apart from newly introduced parameters (\ie, \CA layers, Q-Former~\cite{li2023blip2}) which are always trained at a base learning rate of $10^{-4}$, previously existing parameters (\ie, the image encoder and pretrained language model) are either frozen or trained with a learning rate of $10^{-5}$. For all experiments (except smaller-scale ablations), we train on 64 NVIDIA H100 GPUs with 2 gradient accumulation steps, \ie the maximum number of tokens per gradient update is $128\times \#\text{\{Image}+\text{text tokens\}}$.
}
\vspace{-.15cm}
\label{tab:supp-config}
\end{table*}

%% file: tables/supp_resolution.tex
\begin{table*}[t]
\centering
\footnotesize
\scalebox{.98}{
\setlength{\tabcolsep}{4pt}
\begin{tabular}{lccccccccccc}
\toprule
&  && \multicolumn{3}{c}{High-res Document/Chart Understanding} & \multicolumn{2}{c}{Scene Text Understanding}  & \multicolumn{4}{c}{Knowledge / General QA} \\
\cmidrule(lr){4-6} \cmidrule(lr){7-8} \cmidrule(lr){9-12} & & & \sc{ChartQA} & \sc{DocVQA} & \sc{InfoVQA} & \sc{OCRBench} & \sc{TextVQA} & \sc{RealWorldQA} & \sc{AI2D} & \sc{GQA} & \sc{MME} \\
\midrule
\multirow{4}{*}{Insertion$_\text{He-2B}$} & 504 && 78.0  &  75.2  &  44.0  &  682  &  69.6  &  54.9  &  64.5  &  52.1  & 1686\\
 & 728 && 78.9 &  85.2  & 52.3  &  699  &  72.6  &  57.5  &  65.1  &  52.1  &  1690  \\
 & 952 && 80.0 & 87.2 & 55.3 & 745 & 74.4 & 59.6  & 64.6 & 54.0  &  1676\\
 & 1400 &&  79.6  &  87.8  &  58.8  &  744  &  74.5  &  54.2  &  64.5  &  54.1  &  1679  \\
\midrule
\multirow{4}{*}{\CAHe} & 504 && 72.1  &  69.5  &  38.6  &  672  &  65.9  &  53.2  &  66.5  &  54.0  &  1735  \\
 & 728 && 75.9 &  83.5 &  49.2  &  712  &  72.7  &  58.2  &  66.2  &  54.3  &  1721  \\ 
 & 952 && 75.9  &  85.8  &  51.8  &  722  &  73.8  &  58.0  &  66.2  &  54.3  &  1731   \\
 & 1400 && 75.9 &  & 53.2 & 726 & 74.1 & 57.6 & 66.1 & 54.3 & 1716 \\
\bottomrule
\end{tabular}
}
\caption{\textbf{Impact of image resolution at evaluation.} We evaluate our \CAtv-Helium-2B and Insertion$_\text{He-2B}$ models, for different input image resolutions. For comparison, in \hyperref[tab:sota-images]{Table \ref{tab:sota-images}}, these models are trained and  evaluated at the same max resolution of $952^2$ pixels}
\vspace{-.2cm}
\label{tab:supp-resolution}
\end{table*}

%% file: figs/streaming_supp.tex
\begin{figure*}
    \centering
    \hspace{-.5cm}
    \foreach \FILE in {ACOSBKEW,AGGVGAMC,a\_day\_with\_gymshark} {
    \begin{minipage}{\textwidth}
    \includegraphics[
        trim=1.5em 0 0 0,
        clip,
        width=\textwidth
    ]{figs/video_cc/\FILE.pdf}
    \vspace{-1.4cm}
\input{figs/video_cc/lastqwen_2000_fps2/fps=2.0_gf=1_rp=1.5_t=0.4/\FILE}
\end{minipage}
\vspace{1em}
}
    \caption{\textbf{Live captioning examples -- CA}. Similar to \hyperref[fig:qualitative]{Figure \ref{fig:qualitative}} we show excerpts of video generated by a \CAtv model finetuned on LiveWhisperX. For full video samples, see the \href{kyutai.org/casa}{project page}.
    }
    \vspace{-0.2cm}
    \label{fig:qualitative3}
\end{figure*}

%% file: main.bbl
\begin{thebibliography}{60}
\providecommand{\natexlab}[1]{#1}
\providecommand{\url}[1]{\texttt{#1}}
\expandafter\ifx\csname urlstyle\endcsname\relax
  \providecommand{\doi}[1]{doi: #1}\else
  \providecommand{\doi}{doi: \begingroup \urlstyle{rm}\Url}\fi

\bibitem[Alayrac et~al.(2022)Alayrac, Donahue, Luc, Miech, Barr, Hasson, Lenc, Mensch, Millican, Reynolds, et~al.]{alayrac2022flamingo}
Jean-Baptiste Alayrac, Jeff Donahue, Pauline Luc, Antoine Miech, Iain Barr, Yana Hasson, Karel Lenc, Arthur Mensch, Katherine Millican, Malcolm Reynolds, et~al.
\newblock {Flamingo: a Visual Language Model for Few-Shot Learning}.
\newblock \emph{NeurIPS}, 2022.

\bibitem[An et~al.(2025)An, Xie, Yang, Zhang, Zhao, Cheng, Wang, Xu, Chen, Wu, et~al.]{an2025llavaov15}
Xiang An, Yin Xie, Kaicheng Yang, Wenkang Zhang, Xiuwei Zhao, Zheng Cheng, Yirui Wang, Songcen Xu, Changrui Chen, Chunsheng Wu, et~al.
\newblock Llava-onevision-1.5: Fully open framework for democratized multimodal training.
\newblock \emph{arXiv:2509.23661}, 2025.

\bibitem[Awadalla et~al.(2023)Awadalla, Gao, Gardner, Hessel, Hanafy, Zhu, Marathe, Bitton, Gadre, Sagawa, Jitsev, Kornblith, Koh, Ilharco, Wortsman, and Schmidt]{awadalla2023openflamingo}
Anas Awadalla, Irena Gao, Josh Gardner, Jack Hessel, Yusuf Hanafy, Wanrong Zhu, Kalyani Marathe, Yonatan Bitton, Samir Gadre, Shiori Sagawa, Jenia Jitsev, Simon Kornblith, Pang~Wei Koh, Gabriel Ilharco, Mitchell Wortsman, and Ludwig Schmidt.
\newblock {OpenFlamingo: An Open-Source Framework for Training Large Autoregressive Vision-Language Models}.
\newblock \emph{arXiv:2308.01390}, 2023.

\bibitem[Bai et~al.(2025)Bai, Chen, Liu, Wang, Ge, Song, Dang, Wang, Wang, Tang, et~al.]{bai2025qwen25vl}
Shuai Bai, Keqin Chen, Xuejing Liu, Jialin Wang, Wenbin Ge, Sibo Song, Kai Dang, Peng Wang, Shijie Wang, Jun Tang, et~al.
\newblock {Qwen2.5-VL Technical Report}.
\newblock \emph{arXiv:2502.13923}, 2025.

\bibitem[Chen et~al.(2025)Chen, Zeng, Lin, Li, Ma, and Shou]{chen2025livecc}
Joya Chen, Ziyun Zeng, Yiqi Lin, Wei Li, Zejun Ma, and Mike~Zheng Shou.
\newblock {LiveCC: Learning Video LLM with Streaming Speech Transcription at Scale}.
\newblock In \emph{CVPR}, 2025.

\bibitem[Chen et~al.(2024{\natexlab{a}})Chen, Shen, Zhong, Zhong, Xia, Xu, Yuan, Hu, Wen, Zhang, et~al.]{chen2024evlm}
Kaibing Chen, Dong Shen, Hanwen Zhong, Huasong Zhong, Kui Xia, Di Xu, Wei Yuan, Yifei Hu, Bin Wen, Tianke Zhang, et~al.
\newblock {EVLM: An Efficient Vision-Language Model for Visual Understanding}.
\newblock \emph{arXiv:2407.14177}, 2024{\natexlab{a}}.

\bibitem[Chen et~al.(2024{\natexlab{b}})Chen, Zhao, Liu, Bai, Lin, Zhou, and Chang]{chen2024fastv}
Liang Chen, Haozhe Zhao, Tianyu Liu, Shuai Bai, Junyang Lin, Chang Zhou, and Baobao Chang.
\newblock {An Image is Worth 1/2 Tokens After Layer 2: Plug-and-Play Inference Acceleration for Large Vision-Language Models}.
\newblock In \emph{ECCV}, 2024{\natexlab{b}}.

\bibitem[Chen et~al.(2024{\natexlab{c}})Chen, Wang, Cao, Liu, Gao, Cui, Zhu, Ye, Tian, Liu, et~al.]{chen2024internvl25}
Zhe Chen, Weiyun Wang, Yue Cao, Yangzhou Liu, Zhangwei Gao, Erfei Cui, Jinguo Zhu, Shenglong Ye, Hao Tian, Zhaoyang Liu, et~al.
\newblock {Expanding Performance Boundaries of Open-Source Multimodal Models with Model, Data, and Test-Time Scaling}.
\newblock \emph{arXiv:2412.05271}, 2024{\natexlab{c}}.

\bibitem[Dao(2024)]{dao2023flashattention2}
Tri Dao.
\newblock {Flash{A}ttention-2: Faster Attention with Better Parallelism and Work Partitioning}.
\newblock In \emph{ICLR}, 2024.

\bibitem[Deitke et~al.(2025)Deitke, Clark, Lee, Tripathi, Yang, Park, Salehi, Muennighoff, Lo, Soldaini, et~al.]{deitke2025molmo}
Matt Deitke, Christopher Clark, Sangho Lee, Rohun Tripathi, Yue Yang, Jae~Sung Park, Mohammadreza Salehi, Niklas Muennighoff, Kyle Lo, Luca Soldaini, et~al.
\newblock {Molmo and PixMo: Open Weights and Open Data for State-of-the-Art Vision-Language Models}.
\newblock In \emph{CVPR}, 2025.

\bibitem[Dubey et~al.(2024)Dubey, Jauhri, Pandey, Kadian, Al-Dahle, Letman, Mathur, Schelten, Yang, Fan, et~al.]{dubey2024llama3}
Abhimanyu Dubey, Abhinav Jauhri, Abhinav Pandey, Abhishek Kadian, Ahmad Al-Dahle, Aiesha Letman, Akhil Mathur, Alan Schelten, Amy Yang, Angela Fan, et~al.
\newblock The llama 3 herd of models.
\newblock \emph{2407.21783}, 2024.

\bibitem[Fu et~al.(2023)Fu, Chen, Shen, Qin, Zhang, Lin, Qiu, Lin, Yang, Zheng, Li, Sun, and Ji]{Fu2023MMEAC}
Chaoyou Fu, Peixian Chen, Yunhang Shen, Yulei Qin, Mengdan Zhang, Xu Lin, Zhenyu Qiu, Wei Lin, Jinrui Yang, Xiawu Zheng, Ke Li, Xing Sun, and Rongrong Ji.
\newblock Mme: A comprehensive evaluation benchmark for multimodal large language models.
\newblock \emph{ArXiv}, abs/2306.13394, 2023.

\bibitem[Fu et~al.(2025)Fu, Dai, Luo, Li, Ren, Zhang, Wang, Zhou, Shen, Zhang, et~al.]{fu2025videomme}
Chaoyou Fu, Yuhan Dai, Yongdong Luo, Lei Li, Shuhuai Ren, Renrui Zhang, Zihan Wang, Chenyu Zhou, Yunhang Shen, Mengdan Zhang, et~al.
\newblock {Video-MME: The First-Ever Comprehensive Evaluation Benchmark of Multi-modal LLMs in Video Analysis}.
\newblock In \emph{CVPR}, 2025.

\bibitem[Ge et~al.(2024)Ge, Chen, Lin, Zhu, Liu, Dai, and Zhu]{Ge2024V2PEIM}
Junqi Ge, Ziyi Chen, Jintao Lin, Jinguo Zhu, Xihui Liu, Jifeng Dai, and Xizhou Zhu.
\newblock V2pe: Improving multimodal long-context capability of vision-language models with variable visual position encoding.
\newblock \emph{ArXiv}, abs/2412.09616, 2024.

\bibitem[Hudson and Manning(2019)]{hudson2019gqa}
Drew~A Hudson and Christopher~D Manning.
\newblock {GQA: A New Dataset for Real-World Visual Reasoning and Compositional Question Answering}.
\newblock In \emph{CVPR}, 2019.

\bibitem[Kembhavi et~al.(2016)Kembhavi, Salvato, Kolve, Seo, Hajishirzi, and Farhadi]{kembhavi2016ai2d}
Aniruddha Kembhavi, Mike Salvato, Eric Kolve, Minjoon Seo, Hannaneh Hajishirzi, and Ali Farhadi.
\newblock {A Diagram Is Worth A Dozen Images}.
\newblock In \emph{ECCV}, 2016.

\bibitem[Kyutai(2025)]{helium}
Kyutai.
\newblock {Helium1: A modular and multilingual LLM}, 2025.

\bibitem[Lauren{\c{c}}on et~al.(2024)Lauren{\c{c}}on, Marafioti, Sanh, and Tronchon]{laurenccon2024docmatix}
Hugo Lauren{\c{c}}on, Andr{\'e}s Marafioti, Victor Sanh, and L{\'e}o Tronchon.
\newblock Building and better understanding vision-language models: insights and future directions.
\newblock \emph{arXiv:2408.12637}, 2024.

\bibitem[Li et~al.(2025)Li, Zhang, Chen, Wang, Pu, Cahyono, Yang, Li, and Liu]{li2025otter}
Bo Li, Yuanhan Zhang, Liangyu Chen, Jinghao Wang, Fanyi Pu, Joshua~Adrian Cahyono, Jingkang Yang, Chunyuan Li, and Ziwei Liu.
\newblock {Otter: A Multi-Modal Model with In-Context Instruction Tuning}.
\newblock \emph{IEEE TPAMI}, 2025.

\bibitem[Li et~al.(2023)Li, Li, Savarese, and Hoi]{li2023blip2}
Junnan Li, Dongxu Li, Silvio Savarese, and Steven Hoi.
\newblock {BLIP-2: Bootstrapping Language-Image Pre-training with Frozen Image Encoders and Large Language Models}.
\newblock In \emph{International Conference on Machine Learning (ICML)}, 2023.

\bibitem[Li et~al.(2024{\natexlab{a}})Li, Wang, He, Li, Wang, Liu, Wang, Xu, Chen, Luo, et~al.]{li2024mvbench}
Kunchang Li, Yali Wang, Yinan He, Yizhuo Li, Yi Wang, Yi Liu, Zun Wang, Jilan Xu, Guo Chen, Ping Luo, et~al.
\newblock {MVBench: A Comprehensive Multi-modal Video Understanding Benchmark}.
\newblock In \emph{CVPR}, 2024{\natexlab{a}}.

\bibitem[Li et~al.(2024{\natexlab{b}})Li, Wang, Yu, Zeng, Zhu, Huang, Gao, Li, He, Wang, Qiao, Wang, and Wang]{li2024videochatflash}
Xinhao Li, Yi Wang, Jiashuo Yu, Xiangyu Zeng, Yuhan Zhu, Haian Huang, Jianfei Gao, Kunchang Li, Yinan He, Chenting Wang, Yu Qiao, Yali Wang, and Limin Wang.
\newblock {VideoChat-Flash: Hierarchical Compression for Long-Context Video Modeling}.
\newblock \emph{arXiv:2501.00574}, 2024{\natexlab{b}}.

\bibitem[Liu et~al.(2024{\natexlab{a}})Liu, Yu, Lan, Wang, Fang, Kautz, Li, and Alvare]{liu2024streamchat}
Jihao Liu, Zhiding Yu, Shiyi Lan, Shihao Wang, Rongyao Fang, Jan Kautz, Hongsheng Li, and Jose~M Alvare.
\newblock {StreamChat: Chatting with Streaming Video}.
\newblock \emph{arXiv:2412.08646}, 2024{\natexlab{a}}.

\bibitem[Liu et~al.(2024{\natexlab{b}})Liu, Li, Huang, Yang, Yu, Li, Yin, Liu, Jin, and Bai]{liu2024ocrbench}
Yuliang Liu, Zhang Li, Mingxin Huang, Biao Yang, Wenwen Yu, Chunyuan Li, Xu-Cheng Yin, Cheng-Lin Liu, Lianwen Jin, and Xiang Bai.
\newblock {OCRBench: On the Hidden Mystery of OCR in Large Multimodal Models}.
\newblock \emph{Science China Information Sciences}, 67\penalty0 (12), 2024{\natexlab{b}}.

\bibitem[Maaz et~al.(2024)Maaz, Rasheed, Khan, and Khan]{maaz2023videochatgpt}
Muhammad Maaz, Hanoona Rasheed, Salman Khan, and Fahad~Shahbaz Khan.
\newblock {Video-ChatGPT: Towards Detailed Video Understanding via Large Vision and Language VideoLLaMA 3: Frontier Multimodal Foundation Models for Image and Video Understanding }.
\newblock In \emph{Proceedings of the Association for Computational Linguistics (ACL)}, 2024.

\bibitem[Marafioti et~al.(2025)Marafioti, Zohar, Farr{\'e}, Noyan, Bakouch, Cuenca, Zakka, Allal, Lozhkov, Tazi, et~al.]{marafioti2025smolvlm}
Andr{\'e}s Marafioti, Orr Zohar, Miquel Farr{\'e}, Merve Noyan, Elie Bakouch, Pedro Cuenca, Cyril Zakka, Loubna~Ben Allal, Anton Lozhkov, Nouamane Tazi, et~al.
\newblock {SmolVLM: Redefining small and efficient multimodal models}.
\newblock \emph{arXiv:2504.05299}, 2025.

\bibitem[Masry et~al.(2022)Masry, Do, Tan, Joty, and Hoque]{masry2022chartqa}
Ahmed Masry, Xuan~Long Do, Jia~Qing Tan, Shafiq Joty, and Enamul Hoque.
\newblock {ChartQA: A Benchmark for Question Answering about Charts with Visual and Logical Reasoning}.
\newblock In \emph{{Findings of the association for computational linguistics: ACL}}, 2022.

\bibitem[Mathew et~al.(2021)Mathew, Karatzas, and Jawahar]{mathew2021docvqa}
Minesh Mathew, Dimosthenis Karatzas, and CV Jawahar.
\newblock {DocVQA: A Dataset for VQA on Document Images}.
\newblock In \emph{WACV}, 2021.

\bibitem[Mathew et~al.(2022)Mathew, Bagal, Tito, Karatzas, Valveny, and Jawahar]{mathew2022infovqa}
Minesh Mathew, Viraj Bagal, Rub{\`e}n Tito, Dimosthenis Karatzas, Ernest Valveny, and CV Jawahar.
\newblock {InfographicVQA}.
\newblock In \emph{WACV}, 2022.

\bibitem[Mu et~al.(2023)Mu, Li, and Goodman]{gist}
Jesse Mu, Xiang~Lisa Li, and Noah~D. Goodman.
\newblock Learning to compress prompts with gist tokens.
\newblock \emph{ArXiv}, abs/2304.08467, 2023.

\bibitem[Nassar et~al.(2025)Nassar, Marafioti, Omenetti, Lysak, Livathinos, Auer, Morin, de~Lima, Kim, Gurbuz, et~al.]{nassar2025smoldocling}
Ahmed Nassar, Andres Marafioti, Matteo Omenetti, Maksym Lysak, Nikolaos Livathinos, Christoph Auer, Lucas Morin, Rafael~Teixeira de Lima, Yusik Kim, A~Said Gurbuz, et~al.
\newblock {SmolDocling: An ultra-compact vision-language model for end-to-end multi-modal document conversion}.
\newblock \emph{arXiv:2503.11576}, 2025.

\bibitem[Ouyang et~al.(2025)Ouyang, Qu, Zhou, Zhu, Zhang, Lin, Wang, Zhao, Jiang, Zhao, et~al.]{ouyang2025omnidocbench}
Linke Ouyang, Yuan Qu, Hongbin Zhou, Jiawei Zhu, Rui Zhang, Qunshu Lin, Bin Wang, Zhiyuan Zhao, Man Jiang, Xiaomeng Zhao, et~al.
\newblock {OmniDocBench: Benchmarking Diverse PDF Document Parsing with Comprehensive Annotation}.
\newblock In \emph{CVPR}, 2025.

\bibitem[Patraucean et~al.(2023)Patraucean, Smaira, Gupta, Recasens, Markeeva, Banarse, Koppula, Malinowski, Yang, Doersch, et~al.]{patraucean2023perception}
Viorica Patraucean, Lucas Smaira, Ankush Gupta, Adria Recasens, Larisa Markeeva, Dylan Banarse, Skanda Koppula, Mateusz Malinowski, Yi Yang, Carl Doersch, et~al.
\newblock {Perception Test: A Diagnostic Benchmark for Multimodal Video Models}.
\newblock In \emph{NeurIPS}, 2023.

\bibitem[Qian et~al.(2024)Qian, Dong, Zhang, Zang, Ding, Lin, and Wang]{qian2024videostreaming}
Rui Qian, Xiaoyi Dong, Pan Zhang, Yuhang Zang, Shuangrui Ding, Dahua Lin, and Jiaqi Wang.
\newblock {Streaming Long Video Understanding with Large Language Models}.
\newblock In \emph{NeurIPS}, 2024.

\bibitem[Royer et~al.(2025)Royer, B{\"o}hle, de~Marmiesse, Mazar{\'e}, Zeghidour, D{\'e}fossez, and P{\'e}rez]{royer2025moshivis}
Am{\'e}lie Royer, Moritz B{\"o}hle, Gabriel de Marmiesse, Laurent Mazar{\'e}, Neil Zeghidour, Alexandre D{\'e}fossez, and Patrick P{\'e}rez.
\newblock {Vision-Speech Models: Teaching Speech Models to Converse about Images}.
\newblock \emph{arXiv:2503.15633}, 2025.

\bibitem[Shi et~al.(2016)Shi, Caballero, Husz{\'a}r, Totz, Aitken, Bishop, Rueckert, and Wang]{shi2016real}
Wenzhe Shi, Jose Caballero, Ferenc Husz{\'a}r, Johannes Totz, Andrew~P Aitken, Rob Bishop, Daniel Rueckert, and Zehan Wang.
\newblock {Real-Time Single Image and Video Super-Resolution Using an Efficient Sub-Pixel Convolutional Neural Network}.
\newblock In \emph{CVPR}, 2016.

\bibitem[Singh et~al.(2019)Singh, Natarjan, Shah, Jiang, Chen, Parikh, and Rohrbach]{singh2019textvqa}
Amanpreet Singh, Vivek Natarjan, Meet Shah, Yu Jiang, Xinlei Chen, Devi Parikh, and Marcus Rohrbach.
\newblock {Towards VQA Models That Can Read}.
\newblock In \emph{CVPR}, 2019.

\bibitem[Srinivasan et~al.(2021)Srinivasan, Raman, Chen, Bendersky, and Najork]{srinivasan2021wit}
Krishna Srinivasan, Karthik Raman, Jiecao Chen, Michael Bendersky, and Marc Najork.
\newblock {WIT: Wikipedia-based Image Text Dataset for Multimodal Multilingual Machine Learning}.
\newblock In \emph{Proceedings of the 44th international ACM SIGIR conference on research and development in information retrieval}, 2021.

\bibitem[Tan and Bansal(2019)]{Tan2019LXMERTLC}
Hao~Hao Tan and Mohit Bansal.
\newblock Lxmert: Learning cross-modality encoder representations from transformers.
\newblock In \emph{Conference on Empirical Methods in Natural Language Processing}, 2019.

\bibitem[Vaswani et~al.(2017)Vaswani, Shazeer, Parmar, Uszkoreit, Jones, Gomez, Kaiser, and Polosukhin]{vaswani2017attention}
Ashish Vaswani, Noam Shazeer, Niki Parmar, Jakob Uszkoreit, Llion Jones, Aidan~N Gomez, {\L}ukasz Kaiser, and Illia Polosukhin.
\newblock {Attention is All you Need}.
\newblock In \emph{NeurIPS}, 2017.

\bibitem[Wang et~al.(2024)Wang, Bai, Tan, Wang, Fan, Bai, Chen, Liu, Wang, Ge, et~al.]{wang2024qwen2vl}
Peng Wang, Shuai Bai, Sinan Tan, Shijie Wang, Zhihao Fan, Jinze Bai, Keqin Chen, Xuejing Liu, Jialin Wang, Wenbin Ge, et~al.
\newblock {Qwen2-VL: Enhancing Vision-Language Model's Perception of the World at Any Resolution}.
\newblock \emph{arXiv:2409.12191}, 2024.

\bibitem[Wang et~al.(2025)Wang, Li, Yan, He, Yu, Zeng, Wang, Ma, Huang, Gao, et~al.]{wang2025internvideo2}
Yi Wang, Xinhao Li, Ziang Yan, Yinan He, Jiashuo Yu, Xiangyu Zeng, Chenting Wang, Changlian Ma, Haian Huang, Jianfei Gao, et~al.
\newblock {InternVideo2.5: Empowering Video MLLMs with Long and Rich Context Modeling}.
\newblock \emph{arXiv:2501.12386}, 2025.

\bibitem[Wiedmann et~al.(2025)Wiedmann, Zohar, Mahla, Wang, Li, Frere, von Werra, Gosthipaty, and Marafioti]{wiedmann2025finevision}
Luis Wiedmann, Orr Zohar, Amir Mahla, Xiaohan Wang, Rui Li, Thibaud Frere, Leandro von Werra, Aritra~Roy Gosthipaty, and Andrés Marafioti.
\newblock {FineVision: Open Data Is All You Need}, 2025.

\bibitem[Wu et~al.(2024)Wu, Zhuang, and Chen]{voco}
Linshan Wu, Jiaxin Zhuang, and Hao Chen.
\newblock Voco: A simple-yet-effective volume contrastive learning framework for 3d medical image analysis.
\newblock \emph{2024 IEEE/CVF Conference on Computer Vision and Pattern Recognition (CVPR)}, pages 22873--22882, 2024.

\bibitem[xAI(2024)]{realworldqa}
xAI.
\newblock Grok-1.5 vision preview, 2024.
\newblock Dataset obtained at \url{https://huggingface.co/datasets/lmms-lab/RealWorldQA}.

\bibitem[Xiao et~al.(2021)Xiao, Shang, Yao, and Chua]{xiao2021nextqa}
Junbin Xiao, Xindi Shang, Angela Yao, and Tat-Seng Chua.
\newblock Next-qa: Next phase of question-answering to explaining temporal actions.
\newblock In \emph{CVPR}, 2021.

\bibitem[Xu et~al.(2025)Xu, Xiao, Chen, He, Peng, Lu, and Han]{xu2025streamingvlm}
Ruyi Xu, Guangxuan Xiao, Yukang Chen, Liuning He, Kelly Peng, Yao Lu, and Song Han.
\newblock {StreamingVLM: Real-Time Understanding for Infinite Video Streams}.
\newblock \emph{arXiv:2510.09608}, 2025.

\bibitem[Yang et~al.(2025)Yang, Li, Yang, Zhang, Hui, Zheng, Yu, Gao, Huang, Lv, et~al.]{yang2025qwen3}
An Yang, Anfeng Li, Baosong Yang, Beichen Zhang, Binyuan Hui, Bo Zheng, Bowen Yu, Chang Gao, Chengen Huang, Chenxu Lv, et~al.
\newblock Qwen3 technical report.
\newblock \emph{arXiv:2505.09388}, 2025.

\bibitem[Yang et~al.(2024)Yang, Wang, Zhang, Shen, and Kim]{Yang2024ParallelizingLT}
Songlin Yang, Bailin Wang, Yu Zhang, Yikang Shen, and Yoon Kim.
\newblock Parallelizing linear transformers with the delta rule over sequence length.
\newblock \emph{ArXiv}, abs/2406.06484, 2024.

\bibitem[Ye et~al.(2023)Ye, Hu, Xu, Ye, Yan, Xu, Li, Tian, Qian, Zhang, et~al.]{ye2023ureader}
Jiabo Ye, Anwen Hu, Haiyang Xu, Qinghao Ye, Ming Yan, Guohai Xu, Chenliang Li, Junfeng Tian, Qi Qian, Ji Zhang, et~al.
\newblock {UReader: Universal OCR-free Visually-situated Language Understanding with Multimodal Large Language Model}.
\newblock In \emph{Findings of the Association for Computational Linguistics: EMNLP 2023}, 2023.

\bibitem[Ye et~al.(2025)Ye, Xu, Liu, Hu, Yan, Qian, Zhang, Huang, and Zhou]{ye2025mplugowl3}
Jiabo Ye, Haiyang Xu, Haowei Liu, Anwen Hu, Ming Yan, Qi Qian, Ji Zhang, Fei Huang, and Jingren Zhou.
\newblock {mPLUG-Owl3: Towards Long Image-Sequence Understanding in Multi-Modal Large Language Models}.
\newblock In \emph{ICLR}, 2025.

\bibitem[Zhang et~al.(2025{\natexlab{a}})Zhang, Li, Cheng, Hu, Yuan, Chen, Leng, Jiang, Zhang, Li, et~al.]{zhang2025videollama3}
Boqiang Zhang, Kehan Li, Zesen Cheng, Zhiqiang Hu, Yuqian Yuan, Guanzheng Chen, Sicong Leng, Yuming Jiang, Hang Zhang, Xin Li, et~al.
\newblock {VideoLLaMA 3: Frontier Multimodal Foundation Models for Image and Video Understanding}.
\newblock \emph{arXiv:2501.13106}, 2025{\natexlab{a}}.

\bibitem[Zhang et~al.(2023)Zhang, Li, and Bing]{zhang2023videollama}
Hang Zhang, Xin Li, and Lidong Bing.
\newblock {Video-LLaMA: An Instruction-tuned Audio-Visual Language Model for Video Understanding}.
\newblock \emph{arXiv:2306.02858}, 2023.

\bibitem[Zhang et~al.(2025{\natexlab{b}})Zhang, Wang, Tang, Liu, Feng, and Jin]{zhang2025flashvstream}
Haoji Zhang, Yiqin Wang, Yansong Tang, Yong Liu, Jiashi Feng, and Xiaojie Jin.
\newblock {Flash-VStream: Efficient Real-Time Understanding for Long Video Streams}.
\newblock In \emph{ICCV}, 2025{\natexlab{b}}.

\bibitem[Zhang et~al.(2025{\natexlab{c}})Zhang, Li, Zhang, Pu, Cahyono, Hu, Liu, Zhang, Yang, Li, et~al.]{zhang2025lmms}
Kaichen Zhang, Bo Li, Peiyuan Zhang, Fanyi Pu, Joshua~Adrian Cahyono, Kairui Hu, Shuai Liu, Yuanhan Zhang, Jingkang Yang, Chunyuan Li, et~al.
\newblock Lmms-eval: Reality check on the evaluation of large multimodal models.
\newblock In \emph{Findings of the Association for Computational Linguistics: NAACL 2025}, pages 881--916, 2025{\natexlab{c}}.

\bibitem[Zhang et~al.(2024{\natexlab{a}})Zhang, Hu, Xu, Yan, Xu, Jin, Zhang, and Huang]{zhang2024tinychart}
Liang Zhang, Anwen Hu, Haiyang Xu, Ming Yan, Yichen Xu, Qin Jin, Ji Zhang, and Fei Huang.
\newblock {TinyChart: Efficient chart understanding with program-of-thoughts learning and visual token merging}.
\newblock In \emph{Proceedings of the 2024 Conference on Empirical Methods in Natural Language Processing}, pages 1882--1898, 2024{\natexlab{a}}.

\bibitem[Zhang et~al.(2024{\natexlab{b}})Zhang, Dong, Cao, Zang, Qian, Wei, Chen, Li, Niu, Ding, et~al.]{zhang2024internlm}
Pan Zhang, Xiaoyi Dong, Yuhang Cao, Yuhang Zang, Rui Qian, Xilin Wei, Lin Chen, Yifei Li, Junbo Niu, Shuangrui Ding, et~al.
\newblock {InternLM-XComposer2.5-OmniLive: A Comprehensive Multimodal System for Long-term Streaming Video and Audio Interactions}.
\newblock \emph{arXiv:2412.09596}, 2024{\natexlab{b}}.

\bibitem[Zhang et~al.(2024{\natexlab{c}})Zhang, Wu, Li, Li, Ma, Liu, and Li]{zhang2024llavavid178k}
Yuanhan Zhang, Jinming Wu, Wei Li, Bo Li, Zejun Ma, Ziwei Liu, and Chunyuan Li.
\newblock {Video Instruction Tuning With Synthetic Data}, 2024{\natexlab{c}}.

\bibitem[Zhao et~al.(2023)Zhao, Wu, He, and Huang]{zhao2023svit}
Bo Zhao, Boya Wu, Muyang He, and Tiejun Huang.
\newblock {SVIT: Scaling up Visual Instruction Tuning}.
\newblock \emph{arXiv:2307.04087}, 2023.

\bibitem[Zhou et~al.(2025)Zhou, Shu, Zhao, Wu, Liang, Xiao, Qin, Yang, Xiong, Zhang, et~al.]{zhou2024mlvu}
Junjie Zhou, Yan Shu, Bo Zhao, Boya Wu, Zhengyang Liang, Shitao Xiao, Minghao Qin, Xi Yang, Yongping Xiong, Bo Zhang, et~al.
\newblock {MLVU: Benchmarking Multi-task Long Video Understanding}.
\newblock In \emph{Proceedings of the Computer Vision and Pattern Recognition Conference}, pages 13691--13701, 2025.

\end{thebibliography}
